\documentclass[journal,twoside]{IEEEtran}
\usepackage{cite}
\usepackage{amsmath}
\usepackage{algorithmic}
\usepackage{url}
\usepackage{hyperref}
\usepackage{graphicx}
\usepackage{csquotes}
\usepackage{subfigure}
\usepackage{amssymb,bm}
\usepackage{makecell}
\usepackage{multirow}
\graphicspath{ {./images/} }
\usepackage{xspace}
\usepackage{xcolor}

\newcommand{\latinphrase}[1]{\textit{#1}} 
\newcommand{\etal}{\latinphrase{et~al.}\xspace}
\newcommand{\ie}{\latinphrase{i.e.}\xspace}

\newcommand{\eg}{\latinphrase{e.g.}\xspace}
\newcommand{\etc}{\latinphrase{etc.}\xspace}

\begin{document}
\title{Geometric Back-projection Network \\for Point Cloud Classification}

\author{Shi Qiu,
        Saeed Anwar,
        and Nick Barnes% <-this % stops a space
%\thanks{Manuscript received xxxx xx, 2020; revised xxxx xx, 2020.}
\thanks{S. Qiu and S. Anwar are with Data61, CSIRO (The Commonwealth Scientific and Industrial Research Organisation) and Research School of Engineering, Australian National University, Canberra, ACT 2601, Australia. (Email: \{shi.qiu, saeed.anwar\}@data61.csiro.au).}% <-this % stops a space
\thanks{N. Barnes is with School of Computing, Australian National University, Canberra, ACT 2601, Australia. (Email: nick.barnes@anu.edu.au).}% <-this % stops a space
%\thanks{Color versions of one or more of the figures in this paper are available online at \url{http://ieeexplore.ieee.org}.}
% \thanks{Manuscript received April 19, 2005; revised August 26, 2015.}
}

%\markboth{IEEE TRANSACTIONS ON IMAGE PROCESSING,~Vol.~xx, No.~x, xxxx~2020}{Qiu \MakeLowercase{\textit{et al.}}: Geometric Back-projection Network for Point Cloud Classification}
% \markboth{IEEE TRANSACTIONS ON MULTIMEDIA,~Vol.~xx, No.~x, xxxx~202x}{Qiu \MakeLowercase{\textit{et al.}}: Geometric Back-projection Network for Point Cloud Classification}
\maketitle

\begin{abstract}
As the basic task of point cloud analysis, classification is fundamental but always challenging. To address some unsolved problems of existing methods, we propose a network that captures geometric features of point clouds for better representations. To achieve this, on the one hand, we enrich the geometric information of points in low-level 3D space explicitly. On the other hand, we apply CNN-based structures in high-level feature spaces to learn local geometric context implicitly. Specifically, we leverage an idea of error-correcting feedback structure to capture the local features of point clouds comprehensively. Furthermore, an attention module based on channel affinity assists the feature map to avoid possible redundancy by emphasizing its distinct channels. The performance on both synthetic and real-world point clouds datasets demonstrate the superiority and applicability of our network. Comparing with other state-of-the-art methods, our approach balances accuracy and efficiency.
\end{abstract}


\IEEEpeerreviewmaketitle

\section{Introduction}
\label{sec:intro}
\IEEEPARstart{P}{oint} clouds are one of the fundamental representations of 3D data, and widely used for both research and applications in multimedia~\cite{chen2013point, wu2016fast, de2018graph, valsesia2020learning, zhang2020pointhop} because of the development of 3D sensing technology. Generally, 3D point clouds can be collected by scanners~\cite{blais2004review} utilizing physical touch or non-contact measurements: \eg, light, sound, LiDAR \etc Particularly, LiDAR scanners \cite{jaboyedoff2012use} are in service in many fields including agriculture, biology, and robotics, \etc Due to its tremendous contributions, point cloud analysis attracts much interest for research. More importantly, point cloud classification is always a fundamental challenge due to the following properties:
\begin{itemize}
\item \textbf{Necessity.}   In the same way that ImageNet~\cite{deng2009imagenet} is a litmus test, most 3D work verifies basic performance on classification. Particularly, standalone research on point cloud classification ~\cite{yang2018foldingnet,roveri2018network,chen2018sampled,Uy_2019_ICCV, Li_2020_CVPR, Nezhadarya_2020_CVPR} is essential in 3D vision.
\item  \textbf{Practicability.}  Recently 3D applications are popular and strongly rely on classification. For example, unlocking via face ID requires accurate and efficient 3D classification solutions to identify the users.
\item  \textbf{Difficulty.}  Existing methods face challenges in robustness and efficiency, especially for the rapidly growing needs in the industry. Point cloud classification still requires more efforts to resolve theoretical and practical issues.
\end{itemize}

Traditional algorithms~\cite{schnabel2007efficient,mitra2004registration,rusu2009fast,vosselman20013d} for 3D data usually incorporate geometry estimation and model reconstruction. Assisted by deep learning, recent works on 3D focus on data-driven approaches via Convolutional Neural Networks. Guo \etal~\cite{guo2019deep} categorized CNN approaches to point clouds classification as: multi-view methods (\eg, MVCNN~\cite{su2015multi}), volumetric/mesh methods (\eg, VoxNet~\cite{maturana2015voxnet}), and 3D point methods (\eg, PointNet~\cite{qi2017pointnet}).

Currently, many works explore different methods to improve 3D point cloud processing, but some issues remain unsolved: 1) Besides the 3D coordinates, can we provide more geometric clues for CNN-based feature learning; 2) How can we make the network automatically learn a better representation from the abstract high-level feature space; 3) Moreover, how do we refine the output features to focus on critical information?

State-of-the-art methods~\cite{liu2019relation,wang2017cnn,xu2018spidercnn,li2018so,Hu_2020_CVPR} show that CNN-based feature learning in high-level space perform better when the additional low-level 3D relations are involved. Particularly, \cite{liu2019relation, Hu_2020_CVPR} encode both 3D and high-level features in each layer of their networks. However, repeating similar or incorporating ineffective 3D information in all embedding scales seems redundant and computationally expensive, especially for regular cascaded architecture. Unlike these, we only provide the 3D geometric cues at the beginning, serving as the prior knowledge for our network. To maximize the advantage, such low-level geometric clues should be carefully formed with physically explicit relations to enrich the geometric information for later processing in high-level spaces.

To regulate CNN-based feature learning in embedding space, we present a novel attentional back-projection module capturing an idea of error-correcting feedback structure for point clouds via the incorporation of local geometric context. As supported by substantial biological evidence~\cite{weiss1987dynamic,kitano2002computational}, the feedback mechanism allows modification of the original output via a response to it, which can guide visual tasks for relevant results. Some early work manages to involve the idea of error-feedback in CNNs: \eg, 2D human pose estimation~\cite{carreira2016human}, image super-resolution (SR)~\cite{haris2018deep,liu2019hierarchical}, point cloud generation (3D-SR)~\cite{Li_2019_ICCV}, \etc In order to coordinate the error-feedback idea with complex CNN frameworks, the solutions are generally in two tracks: as in~\cite{carreira2016human}, the feedback loop is realized by minimizing the additional error loss during the back-propagation procedure; while~\cite{haris2018deep, liu2019hierarchical, Li_2019_ICCV} choose to correct the errors by introducing skip-connections in forward-pass. To avoid possible stability problems in training, we prefer the latter solution to realize the error-feedback in a concise way. To the best of our knowledge, such attentional error-correcting feedback structures have not been used for the fundamental problem of point cloud feature learning. Our motivation is to automatically complement the output feature map by comparing the difference between the input and the corresponding \emph{restored} input. Leveraging this structure, we enable the network to learn a better representation of point cloud features.

PointNet~\cite{qi2017pointnet} suggested that symmetric functions can deal with unorderedness of regular point cloud data, since the features can be aggregated consistently regardless of the internal order. Our attentional back-projection module may apply an efficient max-pooling function to extract prominent features from each local neighborhood. However, the extracted information is still insufficient for the precise classification task, as max-pooling can only describe an outline while some local details could be omitted. To address this problem, we propose a simple but effective way to learn complementary fine-grained features via a shared fully connected operation over each local neighborhood. By taking advantage of both prominent and fine-grained local features, our attentional back-projection module is supposed to learn comprehensive representations for point clouds.

Regarding the high-level representations for 3D point clouds, another widely applied mechanism: \emph{Attention} can assist the network to put more emphasis on useful information \cite{vaswani2017attention}. Attention modules are used in many 2D visual problems \eg, image segmentation \cite{wang2018non,hu2018squeeze,fu2019dual, Fang_2019_ICCV}, image denoising \cite{Anwar_2019_ICCV}, person re-identification~\cite{Fang_2019_ICCV}, \etc For 3D point clouds, \cite{xie2018attentional, feng2019point, liu2019l2g} leverage the idea of self-attention~\cite{vaswani2017attention} to find significant point-wise features, while recent work~\cite{Hu_2020_CVPR} distributes weights on neighbors for local aggregation. In terms of the classification task, given a limited number of points, every point-wise feature would be informative. In contrast, possible channel-wise redundancy, accumulated from implicit learning in high-level spaces, would cause side effects for the network. In this case, we propose a Channel-wise Affinity Attention module to refine the feature map based on \emph{affinity} between channels.
\begin{figure}
\begin{center}
\includegraphics[width=\columnwidth]{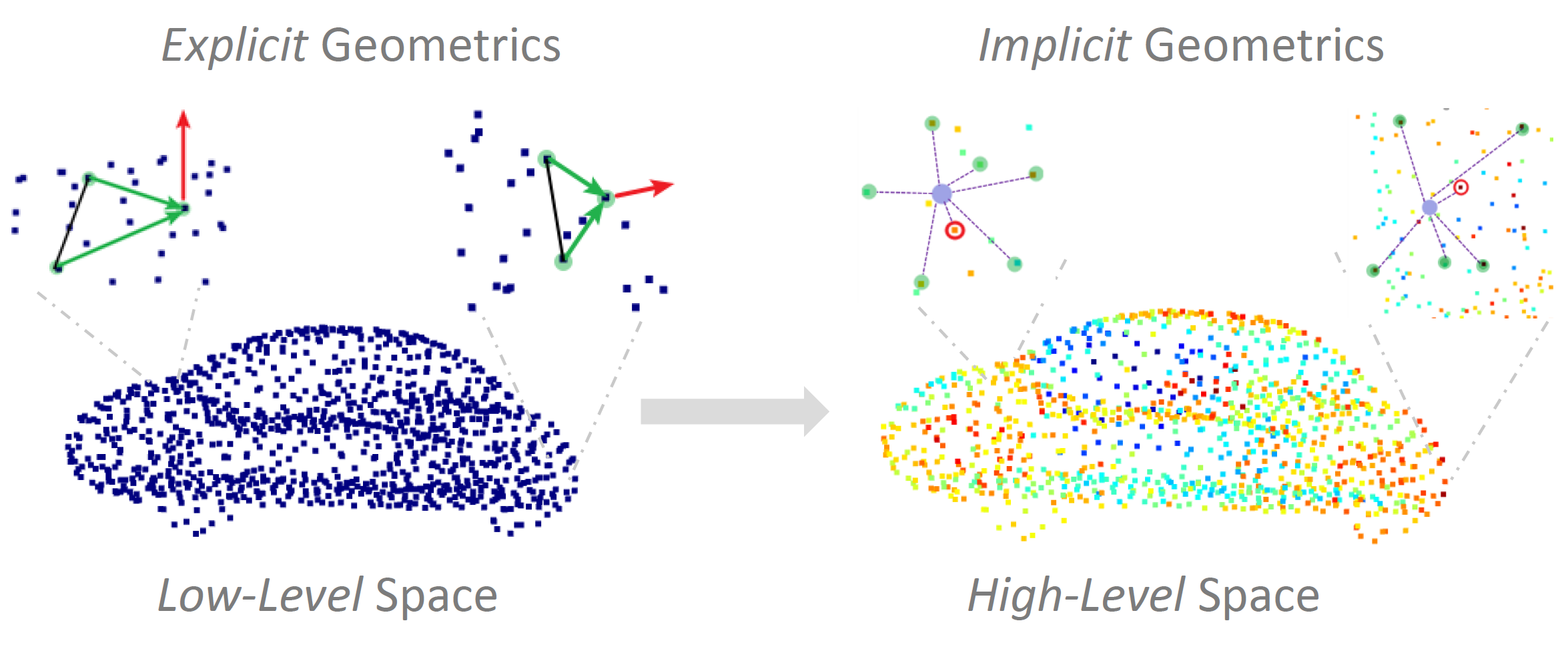}
\end{center}
   \caption{Enriching geometric features. In low-level space, we explicitly estimate geometric items as prior knowledge for the network \eg, edges (green vectors), normals (red vectors). In high-level space, we aggregate neighbors (green points) to implicitly capture both prominent (red points) and fine-grained geometric features (purple points).}
\label{fig:overview}
\end{figure}

The contributions of our work can be summarized as:
\begin{itemize}
\item We propose a CNN network for learning geometric features of point clouds in both low-level and high-level spaces.
\item We design a back-projection CNN module leveraging an idea of error-correcting feedback structure to learn local features of point clouds. 
\item We introduce a channel-wise \emph{affinity} attention module to refine the feature representations of point clouds.
\item We present experimental results showing that our proposed network outperforms state-of-the-art methods on synthetic and real-world 3D point cloud classification benchmarks.
\end{itemize}

The contents of this paper are in the following order: Section~\ref{sec:intro} introduces the paper in general. Section~\ref{sec:rel} reviews previous works on related topics. The details of our approach are explained in Section~\ref{sec:appr}. In Section~\ref{sec:exp}, we present necessary experimental results and ablation studies on different benchmarks. Finally, the conclusions are given in Section~\ref{sec:concl}.

\section{Related Work}
\label{sec:rel}
\subsection{Estimating Geometric Relations}
Although data collection for 3D scattered point cloud is convenient, the main drawback is the lack of geometric information compared with well-constructed mesh or volumetric data. In order to acquire more underlying knowledge of point clouds, conventional methods~\cite{mitra2003estimating, merigot2010voronoi, vosselman20013d} attempted to estimate the geometry of point clouds \ie, face, normals, curvature, \etc In addition, some works proposed different types of hand-crafted features for point cloud recognition or matching: \eg, shape context\cite{kortgen20033d}, point histograms\cite{rusu2009fast}, \etc 

Recently, some CNN-based state-of-the-art methods benefit from permutation invariance of low-level geometry. Xie~\etal~\cite{xie2018attentional} leveraged the idea of 3D geometry estimation, and reproduced shape context calculation using CNN learned features. Yan~\etal~\cite{Yan_2020_CVPR} continually fed the original 3D coordinates into each layer in their network. Further, the encoders in~\cite{liu2019relation, Hu_2020_CVPR} tried to incorporate some 3D relations: \eg, relative positions or distances between the points. Unlike existing solutions, we intend to offer some clues about low-level geometry for the network, instead of repeating similar information for each layer. More explicitly and effectively, we expect that following geometric feature learning in high-level space can benefit from geometric relationships in low-level space.   

\subsection{Learning Local Features}
\begin{figure*}
\begin{center}
\includegraphics[width=0.9\textwidth]{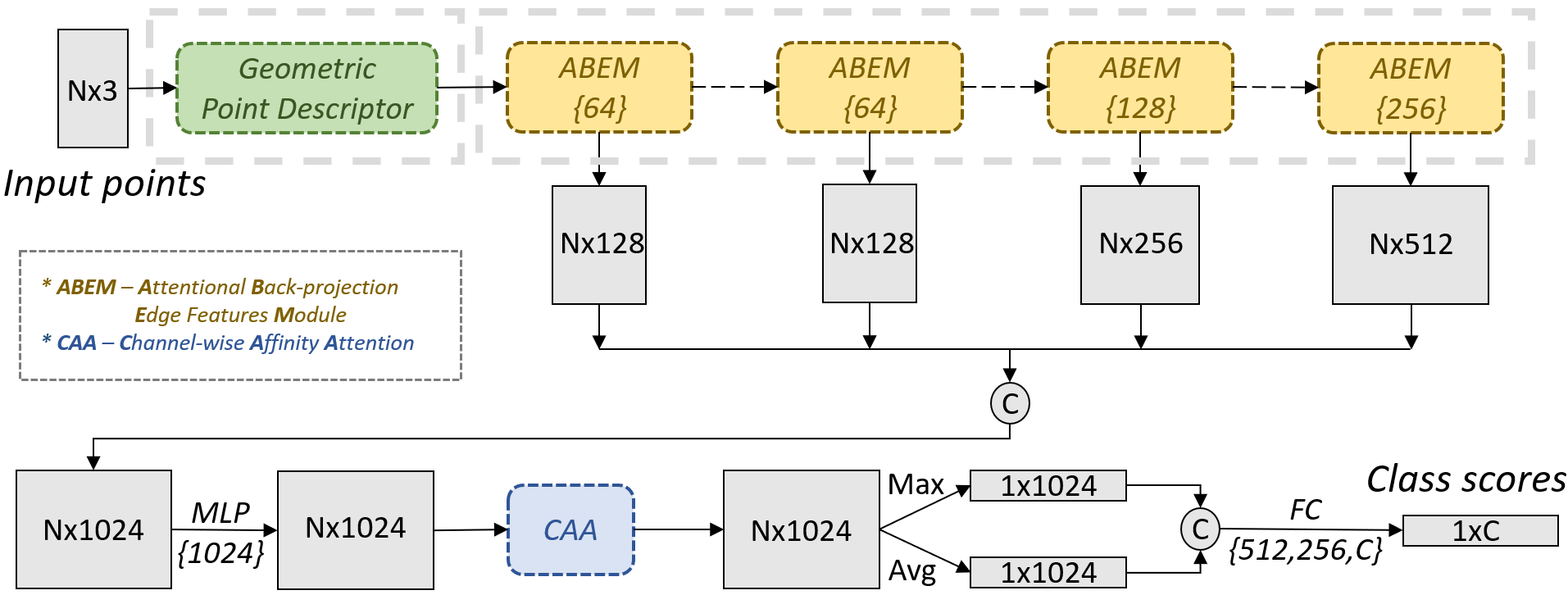}
\end{center}
   \caption{Our network architecture. The Geometric Point Descriptor offers more low-level geometric clues for subsequent high-level geometric feature learning in cascaded ABEMs, representing the point features in multiple scales of embedding space by aggregating local context. The CAA module refines the learned feature map to avoid channel-wise redundancy. Finally, we use the concatenation of max-pooling and average-pooling results, as well as fully connected layers to regress the class scores.}
\label{fig:network}
\end{figure*}

PointNet~\cite{qi2017pointnet} passed the data through multi-layer perceptrons (MLPs) for a high-level feature representation of every single point, and successfully solved the unordered problem of point cloud with a symmetric function. Due to its effectiveness, later works \cite{qi2017pointnet++,li2018pointcnn,wang2019dynamic,liu2019relation} also adopted MLP based operations for point cloud processing. Meanwhile, researchers realized that local features are promising because they involve a more regional context than global features. Although the points are unordered in point cloud data, we may group points based on various metrics. In general, one approach is to select seed points as centroids using a downsampling method (\eg, Farthest Point Sampling in~\cite{qi2017pointnet++, liu2019relation, Yan_2020_CVPR} or Random Sampling in~\cite{Hu_2020_CVPR}), and then apply a query algorithm (\eg, Ball Query in~\cite{qi2017pointnet++} or k-nearest neighbors in~\cite{Hu_2020_CVPR}) based on the 3D Euclidean distance to group points for local clusters. By processing the clusters, the feature of each centroid may represent more local context. 

Another track finds each point's neighbors in embedding space based on N-dimensional Euclidean distance, and then groups them in the form of a high dimensional graph, \eg, as in~\cite{wang2019dynamic, engelmann2019dilated}. In contrast to previous methods, this approach can update the local clusters dynamically on different scales. In terms of feature aggregation, max-pooling~\cite{qi2017pointnet} is widely employed since it can efficiently gather the representative information for an unordered point set. In spite of the benefits, there are some weaknesses to this approach, such as it may involve bias and lose local details. To overcome these issues, we apply an error-correcting feedback structure to regulate the feature learning to reduce possible bias. Moreover, additional fine-grained features aggregated from local neighbors can contribute to more comprehensive feature representations.

\subsection{Attention Mechanism for CNN}
The idea of attention has been successfully used in many areas of Artificial Intelligence. As with human beings, the computational resources of machines are also limited; thus, we need to focus on essential aspects. Previously, Vaswani~\etal~\cite{vaswani2017attention} proposed different types of attention mechanisms for neural machine translation. Subsequently, attention was incorporated in visual tasks: Wang~\etal~\cite{wang2018non} extended the idea of \emph{self-attention} in the spatial domain. Also, SENet~\cite{hu2018squeeze} credits winning the ImageNet~\cite{russakovsky2015imagenet} challenge to its channel-wise attention module. Other works~\cite{woo2018cbam,fu2019dual,li2018harmonious} derive benefits from both spatial and channel domains of 2D images.

In terms of 3D point clouds, attention modules contribute to point cloud approaches for detection~\cite{paigwar2019attentional}, generation~\cite{sun2018pointgrow}, segmentation~\cite{zhang2019pcan,liu2019point2sequence,zhiheng2019pyramnet}, \etc However, there has been limited work in well-designed attention mechanisms targeting 3D point clouds classification. On this front, Xie~\etal~\cite{xie2018attentional} utilized a spatial self-attention module for the learned features based on shape context. Similarly,~\cite{feng2019point, liu2019l2g} intended to enhance the point-wise features by comparing the similarities between points. Subsequent works~\cite{wang2019graph,chen2019gapnet} applied a \emph{Graph Attention} \cite{velivckovic2017graph} module for the constructed graph features on point clouds, while \cite{Hu_2020_CVPR} re-weighted the neighbors for local feature aggregation. Point cloud classification is challenging since a limited number of points are discretely distributed in an unlimited 3D space. Therefore, each point would be informative for the representation of the whole set. Unlike other methods assigning varying weights point-wisely, we attempt to enhance the high-level representation of point cloud by capturing the long-range dependencies along its channels.

\section{Approach}
\label{sec:appr}
In general, our approach is a point-based method. As stated in Section~\ref{sec:intro}, there exists some unsolved problems with the current methods. To tackle the challenges, we start from classical MLP mapping~\cite{qi2017pointnet} and edge features~\cite{wang2019dynamic}. Specifically, MLP performs as a channel-wise fully-connected layer, which is shared among the points in a point cloud feature map. In practice, this operation can be implemented as a 1$\times$1 convolution operating on the point cloud feature map, followed by a batch normalization layer and an activation function:
\begin{equation}
\mathcal{M}(\cdot) := \tau\Big(\bm{BN}\big({c}_{1\times 1}(\cdot)\big)\Big),
\label{equ:mlp}
\end{equation}
where $\mathcal{M}$ is the MLP, $\tau$ is the activation function, $\bm{BN}$ is batch normalization, $c$ is convolution and its subscript presents the filter size.

Besides, edge features ${f_\psi}$ are considered as a type of implicit geometric feature in embedding space since they are crafted following geometric relations between points based on the constraint of high-dimensional Euclidean distance. Specifically, we form the edge features by concatenating the point feature $x_i$ with its edge vectors $x_j-x_i$:
\begin{equation}
{f_\psi}(x_i) = [x_i, x_j-x_i], \quad x_i\in\mathbb{R}^d;\; \quad {\forall}x_j\in Ni(x_i).
\label{equ:edgefeature}
\end{equation}
The $x_i$ means a point feature in $d$-dimensional embedding space, where the entire point cloud feature map is $\mathcal{X}_{{N}\times{d}}$. The $x_j$'s are the local neighbors of $x_i$ found by the k-nearest neighbors ($knn$) algorithm, and \enquote{[\;\;]} denotes a concatenation operation.

Particularly, MLP can encode the local context given by the edge features, denoted as EdgeConv ($\mathcal{E}$):
\begin{equation}
    \mathcal{E}(\cdot) := \mathcal{M}\big({f_\psi}(\cdot)\big),
    \label{equ:edgeconv}
\end{equation}
Physically, EdgeConv crafts radial local-graphs consisting of edges pointing from the neighbors to the centroids, and then mapped by a shared MLP in feature space. With the symmetric function being applied afterward, corresponding regional information will be encoded in each point feature.

As reviewed in Section~\ref{sec:rel}, symmetric functions like max-pooling can summarize an outline of a region by simply extracting the prominent characters. This process is efficient and effective, especially for a large set of points. However, for a relatively small local area with fewer points, we still need a more reasonable way that takes all neighbors into account. In this case, we may intuitively aggregate the local details by calculating a weighted sum of neighboring features. In order to perform the local aggregation adaptively, we can learn the weights through a shared local fully-connected (LFC) layer. Different from an MLP that mainly encodes a single point feature using a $1\times 1$ convolution, an LFC acts as a $1\times k$ convolution over a local area, learns the weights for $k$ neighbors in this area, and finally aggregates the weighted sum of $k$ neighbors' features as the detailed local context.
\begin{equation}
{\mathcal{L}}(\cdot) := \tau\Big(\bm{BN}\big({c}_{1\times k}(\cdot)\big)\Big),
\label{equ:lfc}
\end{equation}
where $\mathcal{L}$ stands for an LFC layer. Therefore, the edge features processed by LFC is denoted as EdgeLFC ($\mathcal{J}$):
\begin{equation}
    \mathcal{J}(\cdot) := \mathcal{L}\big({f_\psi}(\cdot)\big).
\label{equ:edgelfc}
\end{equation}

In general, EdgeLFC can aggregate the local context and map geometric features simultaneously via learnable weights, by which the details of a local area will be retained.

Based on the motivation of enriching geometric features for the point cloud, we intend to provide more low-level geometric clues for following high-level geometric feature learning. As illustrated in Figure~\ref{fig:network}, the network consists of a series of modules learning local context-based feature maps from different scales of embedding space. In this section, we introduce the structures of critical modules and mathematically formulate the relevant operations.

\subsection{Geometric Point Descriptor}
Generally, our geometric network works under a cooperative mechanism between low-level and high-level geometric features. Specifically, the Geometric Point Descriptor module aims to provide information about explicit geometric relations in low-level space for the following implicit geometric feature learning in high-level space. As regular scattered point clouds give 3D coordinates only, we may reinvent additional features with respect to 3D geometry.

As introduced in Section~\ref{sec:intro}, triangular faces are used in mesh or volumetric data, since they can illustrate the outline of 3D objects. Although reconstructing the faces is not the topic of this work, some relevant features can be estimated to expand the low-level geometric representation of the 3D point clouds. As long as the estimating procedure performs equally on all points, the generated features can offer reasonable geometric clues for later processing.

\begin{figure}
\begin{center}
\includegraphics[width=\columnwidth]{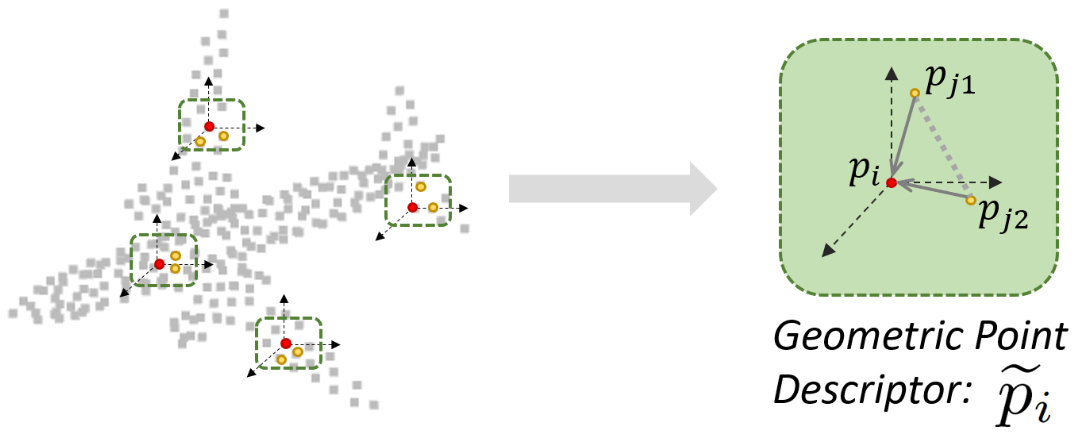}
\end{center}
   \caption{The Geometric Point Descriptor module explicitly enhances the low-level geometric relations between scattered points using triangular features. The red point is an example of a target, while yellow points are corresponding neighbors.}
\label{fig:gp}
\end{figure}
Specifically, we first search the two nearest neighbors of point $p_i\in\mathbb{R}^3$ in 3D space, denoting these as ${p_{j1}}, {p_{j2}}\in\mathbb{R}^3$. By forming a triangle using them, we can explicitly estimate a geometric descriptor, $\widetilde{p}_i$, corresponding to $p_i$:
\begin{equation}
     \widetilde{p}_i = (p_i, normal, {edge}_1, {edge}_2, {{length}_1}, {{length}_2}), \label{equ:nmd}
\end{equation}
concretely:
\begin{small}
$$
\widetilde{p}_i\in\mathbb{R}^{14}
 \begin{cases}
    p_i = (x,y,z); &\mbox{$p_i\in\mathbb{R}^{3}$},\\
    normal = {{edge}_1}\times{{edge}_2}; &\mbox{$normal\in\mathbb{R}^{3}$},\\
    {edge}_1 = {p_{j1}-p_i}; &\mbox{${edge}_1\in\mathbb{R}^{3}$},\\
    {edge}_2 = {p_{j2}-p_i}; &\mbox{${edge}_2\in\mathbb{R}^{3}$},\\
    {length_1} = |{edge}_1|; &\mbox{${length_1}\in\mathbb{R}^{1}$},\\
    {length_2} = |{edge}_2|; &\mbox{${length_2}\in\mathbb{R}^{1}$}.
 \end{cases}
$$
\end{small}

Through this simple process of concise formulation and reduced computation, we expand the low-level geometric features of $p_i$ with estimated edges and a normal vector besides its 3D coordinates. In general, the Geometric Point Descriptor module helps to enhance the scattered point clouds using additional internal geometric relations, and further benefits the following feature learning in high-level space. For instance, the CNN will explicitly recognize an outlier based on the geometric characters of its larger edge lengths, \ie, $5^{th}$ and $6^{th}$ features in Equation~\ref{equ:nmd}. More discussions on this module are in Section~\ref{sec:exp_nmd}.

\subsection{Attentional Back-projection Edge Features Module}
\label{sec:abef}
Besides the low-level geometric information, our network also requires the modules for geometric feature learning in high-level space. Mainly, an error feedback mechanism aims to accurately control a process by monitoring its actual output, feeding the errors, and attaining the desired output. Inspired by this, we can design a structure to regulate the learning process by projecting a restored signal back. Assisted by neural network training mechanism, a specifically defined error signal is expected to constrain feature learning and mitigate possible bias against desired output as training continues.

Here we propose the Attentional Back-projection Edge Features Module (ABEM) to represent the local geometric context for point clouds in embedding space. In general, ABEM consists of two branches encoding local prominent and fine-grained geometric features, respectively.
\begin{figure*}
\begin{center}
\includegraphics[width=0.8\textwidth]{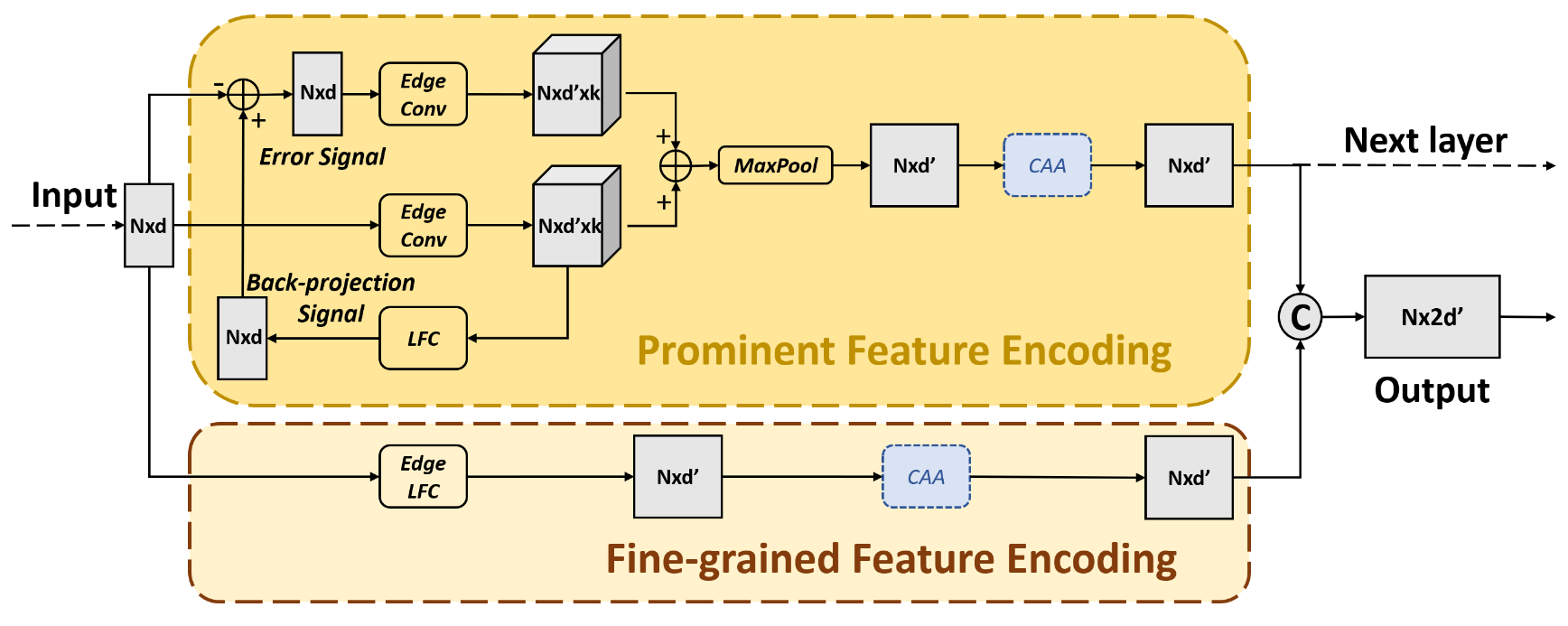}
\end{center}
   \caption{Attentional Back-projection Edge Features Module (ABEM) learns local geometric features in high-level spaces from two branches: one extracts prominent local context, while the other aggregates fine-grained information as a complement. We take the concatenated outputs to form the final feature representations of the point clouds and only pass the prominent features for the next layer of learning. (\emph{Note}: The ABEM works in a forward manner as introduced in Section~\ref{sec:abef}.)}
\label{fig:abem}
\end{figure*}
\subsubsection{Prominent Feature Encoding}
Similar to previous annotations, $x_i\in\mathbb{R}^d$ is a point feature of the feature map $\mathcal{X}_{{N}\times{d}}$. Specifically, we add different subscripts to the operations defined in Equation~\ref{equ:mlp},\ref{equ:edgeconv},\ref{equ:lfc},\ref{equ:edgelfc} if they are applied more than once. 

Firstly, we employ an EdgeConv denoted as $\mathcal{E}_{\Phi}$ for the input feature map since this operation can represent the geometric relations of a local area in high-level space. With Equation~\ref{equ:edgeconv} applied, the output of $\mathcal{E}_{\Phi}$ is:
\begin{equation}
    f_{{\Phi}_i} = \mathcal{E}_{\Phi}(x_i) = {\mathcal{M}_{{\Phi}}}\big({f_\psi}(x_i)\big); \quad \quad\;f_{{\Phi}_i}\in\mathbb{R}^{{d^{\prime}}\times k}, 
    \label{equ:forwardout}
\end{equation}
where $k$ is the number of neighbours of a point.

Following our EdgeConv $\mathcal{E}_{\Phi}$, an ideal output feature map should fully encode local geometric details. On the other hand, if the output $f_{{\Phi}_i}$ is indeed informative, we may restore the original input $x_i$ from $f_{{\Phi}_i}$ through a back-projection. Therefore, such back-projection should simulate the \emph{reverse} process of EdgeConv. In our proposal, we use a shared LFC ($\mathcal{L}_{\Gamma}$) layer to realize the back-projection for $f_{{\Phi}_i}$. In contrast to EdgeConv expanding a center point to local neighbors in the new embedding space, the physical meaning of LFC operation is to pull the neighbors back to the previous space $({d^{\prime}}\rightarrow d)$ and aggregate at a center point $(k\rightarrow 1)$ via learnable weights. In general, we define the restored feature as \emph{Back-projection Signal} ${x_i}^{\prime}$. Based on Equations~\ref{equ:lfc} and~\ref{equ:forwardout}:
\begin{equation}
        {x_i}^{\prime} = {\mathcal{L}_{\Gamma}}(f_{{\Phi}_i}) = {\mathcal{L}_{\Gamma}}\Big({\mathcal{M}_{{\Phi}}}\big({f_\psi}(x_i)\big)\Big); \quad \quad \; {x_i}^{\prime}\in\mathbb{R}^d.
\end{equation}

Further, we take the difference between the original and restored (\ie, \emph{Back-projection Signal}) inputs, $\Delta x_i$, to formulate the corresponding \emph{Error Signal}:
\begin{equation}
    \Delta x_i = {x_i}^{\prime} - x_i; \; \quad \quad
    \Delta x_i\in\mathbb{R}^d,
\label{equ:errorsignal}
\end{equation}
and the impact of the \emph{Error Signal} on $f_{{\Phi}_i}$ can be estimated by another EdgeConv ($\mathcal{E}_{\Upsilon}$) projection:
\begin{equation}
    f_{{\Upsilon}_i} = \mathcal{E}_{\Upsilon}(\Delta x_i) = {\mathcal{M}_{{\Upsilon}}}\big({f_\psi}(\Delta x_i)\big); \quad \quad\;f_{{\Upsilon}_i}\in\mathbb{R}^{{d^{\prime}}\times k}. \label{equ:correct}
\end{equation}

Although this approach implies a similar idea to the auto-encoder of restoring the input~\cite{badrinarayanan2017segnet, qi2017pointnet++}, there are clear differences between them. Auto-encoders usually changes the size of the point cloud and concatenates the features from different resolutions for a comprehensive representation; in our method, the size of the point cloud is constant while the dimension of features is manipulated to estimate lost information (\ie \emph{Error Signal}) of the input. Moreover, different from the auto-encoder concatenating features for complementary information, we can correct the output $f_{{\Phi}_i}$ by adding the \emph{Error Signal} impact, $f_{{\Upsilon}_i}$. As Figure \ref{fig:abem} shows, max-pooling is applied over the local area to extract prominent features, and the Channel-wise Affinity Attention (CAA, see Section \ref{ref:CAA} and Equation \ref{equ:caa} for details) module is placed afterwards in order to refine the final feature representation. In general, the output of the Prominent Feature Encoding branch is:
\begin{equation}
    f_{M_i} = \bm{CAA}\big(\max_{\{k\}} (f_{{\Phi}_i} + f_{{\Upsilon}_i})\big);  \quad \quad \; {f_{M_i}}\in\mathbb{R}^{d^\prime}. \label{equ:pfe}
\end{equation}  
\subsubsection{Fine-grained Feature Encoding}
Although prominent features extracted by max-pooling can efficiently encode much geometric information for simple point clouds, more fine-grained features are required for comprehensive feature representation, especially for some challenging cases \eg, real objects, complex scenes, or similar shapes. Especially, we apply an EdgeLFC (explained in Equation~\ref{equ:edgelfc}), $\mathcal{J}_{\Theta}$, to aggregate the local geometric details from all neighbors, in order to complement the prominent feature encoding outputs. With a CAA module applied, the output of the Fine-grained Feature Encoding branch is:
\begin{equation}
    f_{A_i} = \bm{CAA}\big(\mathcal{J}_{\Theta}(x_i)\big) = \bm{CAA}\Big(\mathcal{L}_\Theta\big({f_\psi}(x_i)\big)\Big);  \quad  {f_{A_i}}\in\mathbb{R}^{d^\prime}. \label{equ:ffe}
\end{equation}

In the end, we concatenate the outputs of the two encoding branches as the learned geometric features by the ABEM:
\begin{equation}
    f_i = \bm{ABEM}\big(x_i\big) = \bm{concat}(f_{M_i}, f_{A_i});  \quad  {f_i}\in\mathbb{R}^{2d^\prime}. \label{equ:abem}
\end{equation}

\subsection{Channel-wise Affinity Attention Module}
\label{ref:CAA}
As explained in Section \ref{sec:rel}, most attention structures regarding point clouds operate in point-space, but the effects are not apparent as work~\cite{xie2018attentional} and Table~\ref{tab:attention} indicate. Instead, distributing attention weights along channels is worth investigating. Inspired by the spacetime non-local block~\cite{wang2018non}, we calculate the long-range dependencies of channels without being concerned by point cloud's unorderedness.  Practically, the corresponding computation usually costs a lot of time and memory. Therefore, we need to search for an effective and efficient way to avoid channel-wise redundancy in an abstract embedding space.

\begin{figure}
\begin{center}
\subfigure[Channel-wise Affinity Attention module (CAA).]{                    
    \includegraphics[width=0.95\columnwidth]{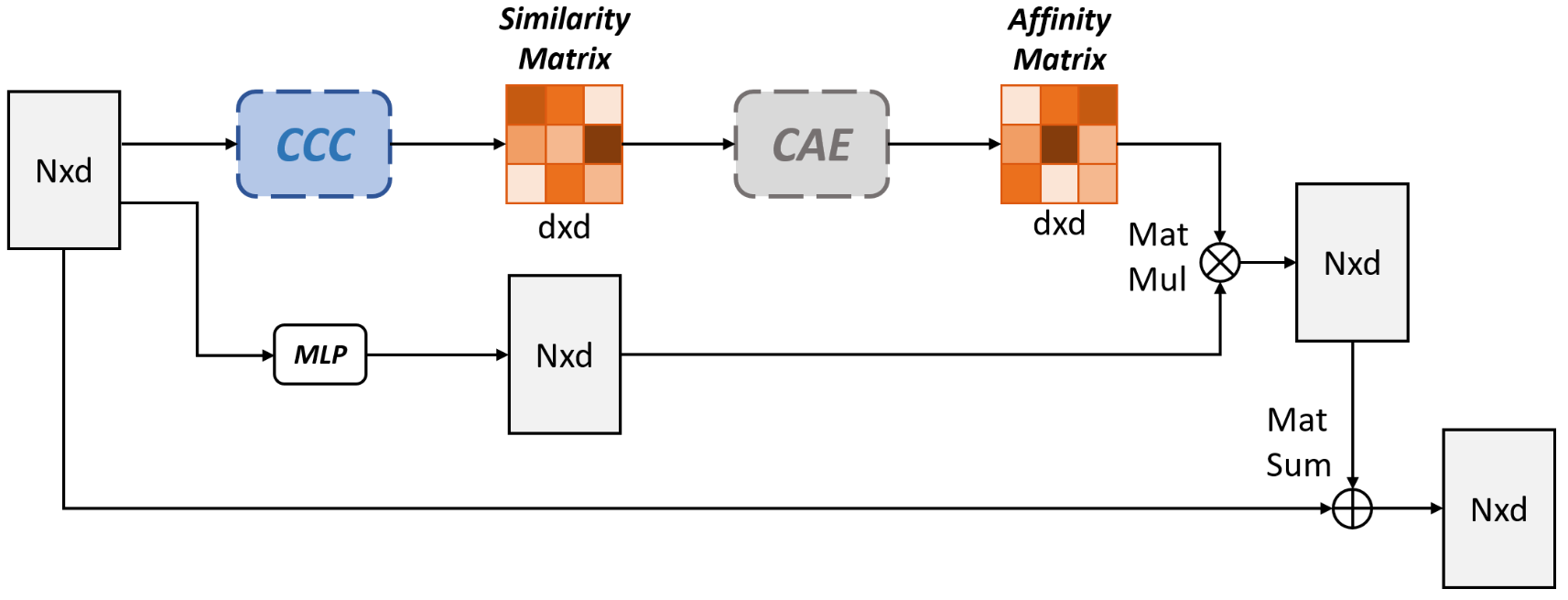}\label{fig:caa}                
    }
\subfigure[Details of Compact Channel-wise Comparator block (CCC) and Channel Affinity Estimator block (CAE).]{                    
    \includegraphics[width=0.98\columnwidth]{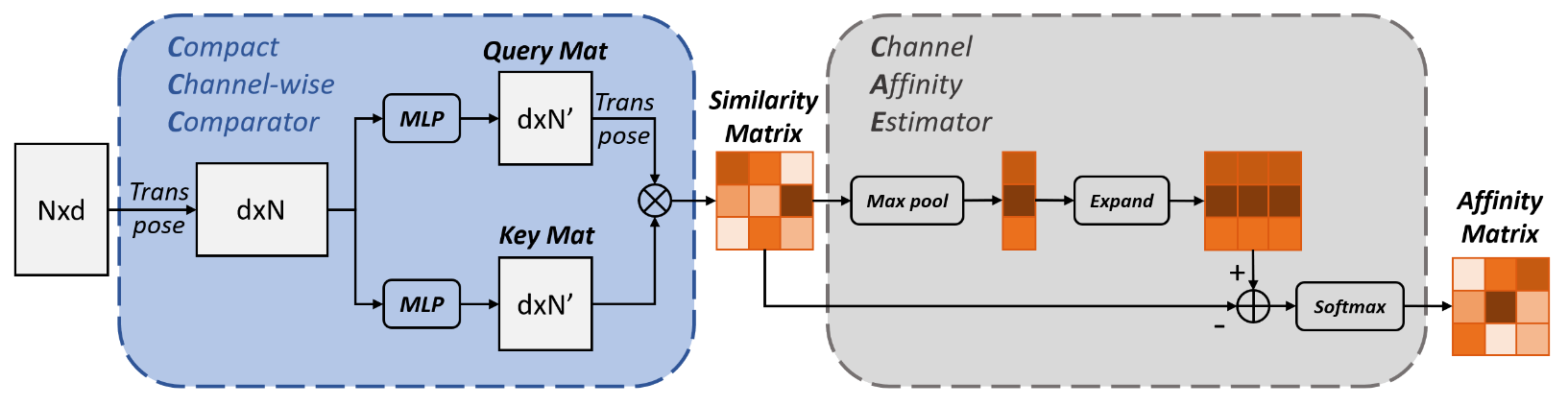}\label{fig:ccccae}                
    }
\end{center}
  \caption{Channel-wise Affinity Attention module (CAA, in~\ref{fig:caa}). Specifically, the Compact Channel-wise Comparator block (CCC, in blue of~\ref{fig:ccccae}) can approximate the similarity matrix between the channels of the input feature map. Then Channel Affinity Estimator (CAE, in the grey of ~\ref{fig:ccccae}) takes the similarity matrix for the calculation of the affinity matrix.}
\label{fig:caa_all}
\end{figure}

We propose our Channel-wise Affinity Attention (CAA) module targeting the channels of high-level point cloud feature maps. As Figure \ref{fig:caa_all} shows, the main structure includes a Compact Channel-wise Comparator (CCC) block, a Channel Affinity Estimator (CAE) block, and a residual connection.

\subsubsection{Compact Channel-wise Comparator block}
Since the CAA module mainly focuses on channel-space, it is necessary to reduce the computing cost caused by the complexity in point-space. In the case of a given $d$-dimensional feature map $\mathcal{F}_{{N}\times{d}}$, the Compact Channel-wise Comparator (CCC) block can simplify context in each channel by an shared MLP operating on channel vector ${c_i}\;(where\; {c_i}\in{\mathbb{R}^N}; and \;{\mathcal{F}_{{N}\times{d}}} = [c_1,\;c_2,\;...\;,\;c_d ]$) to implicitly replace $N$ original points with a smaller number ${N^\prime}={N/ratio};\;ratio>1$. In contrast to taking all points or selecting certain points based on additional metrics, CCC can efficiently reduce the size but sufficiently retain the information of each channel:
$$q_i = \mathcal{M}_q({c_i}); \;{q_i}\in{\mathbb{R}^{N^\prime}},$$
$$k_i = \mathcal{M}_k({c_i}); \;{k_i}\in{\mathbb{R}^{N^\prime}}.$$
Specifically, $\mathcal{M}_q(\cdot)$ and $\mathcal{M}_k(\cdot)$ are two MLPs operating for \emph{Query Matrix} and \emph{Key Matrix} \cite{vaswani2017attention}:
$$\mathcal{Q}_{{N^\prime}\times{d}} = [q_1,\;q_2,\;...\;,\;q_d ],$$
$$\mathcal{K}_{{N^\prime}\times{d}} = [k_1,\;k_2,\;...\;,\;k_d ],$$
and we apply the product of the transposed \emph{Query Matrix} and \emph{Key Matrix} to estimate the corresponding channel-wise \emph{Similarity Matrix}:
$$\mathcal{S}_{{d}\times{d}} = {\mathcal{Q}}^T \mathcal{K},$$
where $\mathcal{S}_{{i},{j}}$ approximates the similarity between the $i^{th}$ channel and the $j^{th}$ channel of the given feature map $\mathcal{F}_{{N}\times{d}}$.

\subsubsection{Channel Affinity Estimator block}
Typical self-attention structures used to calculate the long-range dependencies in spatial data based on inner-products since the values can somehow represent the similarities between the items. In contrast, we define the non-similarities between channels and term it \emph{Channel Affinity}. In our approach, the \emph{Channel Affinity Matrix} of the feature map $\mathcal{F}_{{N}\times{d}}$:
\begin{equation}
    {\mathcal{A}}_{{d}\times{d}} = \bm{softmax}\Big(\underset{{1\rightarrow d}}{\bm{expand}}\big({\max_{d\rightarrow1}(\mathcal{S})}\big) - \mathcal{S}\Big). \label{equ:affinity}
\end{equation}

Particularly, we select the maximum similarities along the columns of $\mathcal{S}$, and then expand them into the same size of $\mathcal{S}$. By subtracting the original $\mathcal{S}$ from the expanded matrix, the channels with higher similarities have lower affinities (illustrated in Figure \ref{fig:ccccae}). Besides, \emph{softmax} is added to normalize the values, since ${\mathcal{A}}_{{d}\times{d}}$ is used as the weight matrix for refinement. In this way, each channel can put higher weights on other distinct ones, thereby avoid aggregating similar/redundant information. 

According to the weight matrix, we can refine each channel of a point feature by taking a weighted sum of all channels. Here, we apply another MLP, $\mathcal{M}_v(\cdot)$, to get the \emph{Value Matrix}:
$$\mathcal{V}_{{N}\times{d}} = [v_1,\;v_2,\;...\;,\;v_d ],$$
$$v_i = \mathcal{M}_v({c_i}); \;{v_i}\in{\mathbb{R}^{N}}.$$
Therefore, such a refining process can be easily realized by multiplying the \emph{Value Matrix} $\mathcal{V}_{{N}\times{d}}$ and the \emph{Channel Affinity Matrix} ${\mathcal{A}}_{{d}\times{d}}$. Additionally, we use a residual connection with a learnable weight $\alpha$ to ease training. The refined feature map by CAA is given:
\begin{equation}
    {{\mathcal{F}}^\prime}_{N\times d} = {\bm{CAA}(\mathcal{F})} = \mathcal{F} + {\alpha}\cdot{\mathcal{V}\mathcal{A}}. \label{equ:caa}
\end{equation}

\section{Experiments}
\label{sec:exp}
In this section, we first provide the experimental settings utilized for evaluation. We then compare the performances of our approach against state-of-the-art methods on synthetic and real-world point clouds. Furthermore, we analyze the effects of different components of our network by conducting relevant ablation studies. By measuring the complexity, visualizing the intermediate outputs, and testing on semantic segmentation task, we comprehensively demonstrate our network's properties.

\begin{figure*}
\begin{center}
\subfigure[Examples of \emph{ModelNet40} dataset.]{                    
    \includegraphics[width=0.93\textwidth]{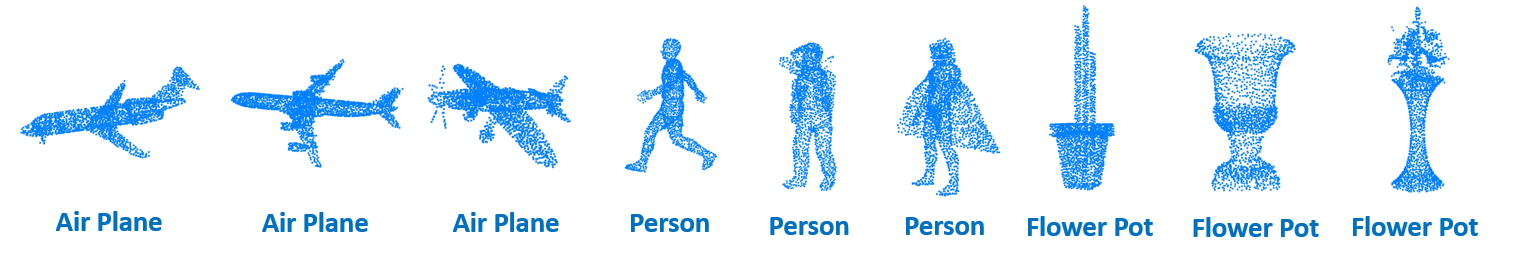}\label{fig:modelnet}                
    }
\subfigure[Examples of \emph{ScanObjectNN} dataset.]{                    
    \includegraphics[width=0.95\textwidth]{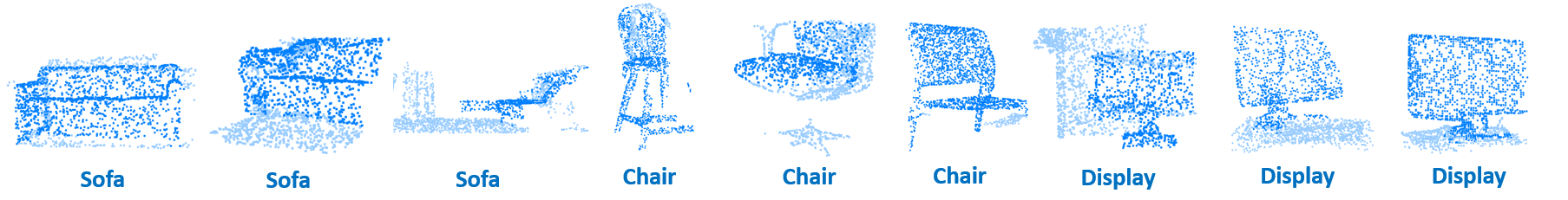}\label{fig:scan}                
    }
\end{center}
  \caption{Examples from point cloud classification datasets. \emph{ModelNet40} include more categories of point clouds as in Figure~\ref{fig:modelnet}, while the samples shown in Figure~\ref{fig:scan} from \emph{ScanObjectNN} are more practically challenging due to complex background (points in lighter color), missing parts, and deformations. }
\label{fig:examples}
\end{figure*}
\subsection{Experimental Settings}
\subsubsection{Implementation}
Our proposed network starts with a Geometric Point Descriptor module, which expands the 3D coordinates of each input point into a 14-degree low-level geometric vector as in Equation~\ref{equ:nmd}. Next, the expanded features are passed through four cascaded Attentional Back-projection Edge Features Modules (ABEMs), aggregating ($k=20$) local neighbors to extract high-level geometric features in different scales (64, 64, 128, and 256) of embedding space. These parameters are similarly adopted from~\cite{wang2019dynamic,liu2019relation,qi2017pointnet,qi2017pointnet++}.

To incorporate the information from different scales, we concatenate the output feature maps of the ABEMs together, and a shared MLP with CAA module can further integrate them into a 1024-dimensional representation. Then we point-wisely apply max-pooling and average-pooling in parallel to form a global embedding vector, by which the other three fully connected layers (having 512, 256, $c$ output) can regress the confidence scores for all possible categories. In the end, we employ cross-entropy between predictions and ground-truth labels as our loss function. This project is implemented with PyTorch, relevant training and testing works are on Linux and GeForce RTX 2080Ti GPUs. \footnote{The codes and models will be available at~\url{https://github.com/ShiQiu0419/GBNet}}

\subsubsection{Training}
We apply Stochastic Gradient Descent (SGD) with the momentum of 0.9 as the optimizer for training, and its initial learning rate of 0.1 decreases to 0.001 by cosine annealing \cite{loshchilov2016sgdr}. The batch size is set to 32 for 300 epochs of training. Besides, training data is augmented with random scaling and translation as in \cite{wang2019dynamic}, while there is no pre or post-processing performed during testing.

 \begin{table}
\begin{center}
\caption{Classification results (\%) on \emph{ModelNet40} benchmark. ($coords$: $(x,y,z)$ coordinates, $norm$: point normal, $voting$: multi-votes evaluation strategy, $k$:$\times 2^{10}$, -: unknown)}
\resizebox{0.95\columnwidth}{!}{
\begin{tabular}{||c|c|c|c|c||}
\hline
method             & input type  & \#points & avg class acc. & overall acc. \\ \hline\hline
ECC \cite{simonovsky2017dynamic}                & $coords$      & $1k$       & 83.2           & 87.4         \\ 
PointNet \cite{qi2017pointnet}           & $coords$      & $1k$       & 86.0           & 89.2         \\ 
SCN \cite{xie2018attentional}                & $coords$      & $1k$       & 87.6           & 90.0         \\ 
Kd-Net \cite{klokov2017escape}             & $coords$      & $1k$       & -              & 90.6         \\ 
PointCNN \cite{li2018pointcnn}           & $coords$      & $1k$       & 88.1           & 92.2         \\
PCNN \cite{atzmon2018point}               & $coords$      & $1k$       & -              & 92.3         \\
DensePoint \cite{Liu_2019_ICCV}         & $coords$      & $1k$       & -    & 92.8         \\
RS-CNN \cite{liu2019relation}           & $coords$      & $1k$       & -              & 92.9         \\ 
DGCNN \cite{wang2019dynamic}              & $coords$      & $1k$       & 90.2           & 92.9         \\ 
KP-Conv \cite{Thomas_2019_ICCV}              & $coords$      & $1k$       & -           & 92.9         \\
PointASNL \cite{Yan_2020_CVPR}              & $coords$      & $1k$       & -           & 92.9         \\
 \hline\hline
\textbf{Ours}   & $\bm{coords}$&$\bm{1k}$&\textbf{91.0}&\textbf{93.8}   \\ \hline\hline
RGCNN \cite{te2018rgcnn}             & $coords + norm$      & $1k$       & 87.3           & 90.5         \\ 
SO-Net \cite{li2018so}             & $coords$      & $2k$       & 87.3           & 90.9         \\ 
PointNet++ \cite{qi2017pointnet++}         & $coords + norm$ & $5k$       & -              & 91.9         \\ 
SpiderCNN \cite{xu2018spidercnn}          & $coords + norm$ & $5k$       & -              & 92.4         \\
DensePoint \cite{Liu_2019_ICCV}         & $coords + voting$      & $1k$       & -    & 93.2         \\
SO-Net \cite{li2018so}             & $coords + norm$ & $5k$       & 90.8           & 93.4         \\
DGCNN \cite{wang2019dynamic}              & $coords$      & $2k$       & 90.7           & 93.5         \\
RS-CNN \cite{liu2019relation} & $coords + voting$      & $1k$       & -              & 93.6         \\\hline
\end{tabular}
\label{tab:modelnet}
}
\end{center}
\end{table}
\begin{table*}
\begin{center}
\caption{Classification results (\%) on \emph{ScanObjectNN} benchmark. }
\resizebox{\textwidth}{!}{
\begin{tabular}{||c|c|c|c c c c c c c c c c c c c c c||}
\hline
& overall acc. & avg class acc. & bag  & bin  & box  & cabinet & chair & desk & display & door & shelf & table & bed  & pillow & sink & sofa & toilet \\ \hline\hline
\# shapes  &      -       &      -         & 298  & 794  & 406  & 1344    & 1585  & 592  & 678     & 892  & 1084  & 922   & 564  & 405    & 469  & 1058 & 325    \\ \hline\hline
3DmFV \cite{ben20183dmfv}      & 63           & 58.1           & 39.8 & 62.8 & 15.0 & 65.1    & 84.4  & 36.0 & 62.3    & 85.2 & 60.6  & 66.7  & 51.8 & 61.9   & 46.7 & 72.4 & 61.2   \\ \hline
PointNet \cite{qi2017pointnet}   & 68.2         & 63.4           & 36.1 & 69.8 & 10.5 & 62.6    & 89.0  & 50.0 & 73.0    & \textbf{93.8} & 72.6  & 67.8  & 61.8 & 67.6   & 64.2 & 76.7 & 55.3   \\ \hline
SpiderCNN \cite{xu2018spidercnn}  & 73.7         & 69.8           & 43.4 & 75.9 & 12.8 & 74.2    & 89.0  & 65.3 & 74.5    & 91.4 & 78.0  & 65.9  & 69.1 & 80.0   & 65.8 & 90.5 & 70.6   \\ \hline
PointNet++ \cite{qi2017pointnet++} & 77.9         & 75.4           & 49.4 & 84.4 & 31.6 & 77.4    & 91.3  & 74.0 & 79.4    & 85.2 & 72.6  & 72.6  & 75.5 & \textbf{81.0}   & \textbf{80.8} & 90.5 & \textbf{85.9}   \\ \hline
DGCNN \cite{wang2019dynamic}      & 78.1         & 73.6           & 49.4 & 82.4 & 33.1 & 83.9    & 91.8  & 63.3 & 77.0    & 89.0 & 79.3  & \textbf{77.4}  & 64.5 & 77.1   & 75.0 & 91.4 & 69.4   \\ \hline
PointCNN \cite{li2018pointcnn}   & 78.5         & 75.1           & 57.8 & 82.9 & 33.1 & 83.6    & \textbf{92.6}  & 65.3 & 78.4    & 84.8 & 84.2  & 67.4  & 80.0 & 80.0   & 72.5 & 91.9 & 71.8   \\ \hline
BGA-DGCNN \cite{Uy_2019_ICCV}	&79.7	&75.7	&48.2	&81.9	&30.1	&\textbf{84.4}	&\textbf{92.6}	&\textbf{77.3}	&80.4	&92.4	&80.5	&74.1	&72.7	&78.1	&79.2	&91.0	&72.9 \\ \hline
BGA-PN++ \cite{Uy_2019_ICCV} &80.2	&77.5	&54.2	&\textbf{85.9}	&39.8	&81.7	&90.8	&76.0	&84.3	&87.6	&78.4	&74.4	&73.6	&80.0	&77.5	&91.9	&\textbf{85.9} \\ \hline
\textbf{Ours}       & \textbf{80.5}         & \textbf{77.8}           & \textbf{59.0} & 84.4 & \textbf{44.4} & 78.2    & 92.1  & 66 & \textbf{91.2}    & 91.0 & \textbf{86.7}  & 70.4  & \textbf{82.7} & 78.1   & 72.5 & \textbf{92.4} & 77.6   \\ \hline\hline
\multicolumn{18}{||c||}{$F_1$ Scores} \\ \hline
Methhod & overall $F_1$ & avg class $F_1$ & bag  & bin  & box  & cabinet & chair & desk & display & door & shelf & table & bed  & pillow & sink & sofa & toilet \\ \hline
DGCNN \cite{wang2019dynamic}  	&0.78 &0.75   &0.58	&0.80	&0.38	&0.76	&0.87	&0.69	&0.84	&0.89	&0.82	&\textbf{0.78}	&0.68	&0.76	&0.79	&0.85	&0.77\\ \hline
PointCNN \cite{li2018pointcnn}   &0.78 &0.77  &0.60	&0.78	&0.45	&\textbf{0.79}	&0.86	&0.68	&0.84	&0.89	&0.79	&0.71	&0.84	&0.75	&\textbf{0.83}	&\textbf{0.88}	&0.82 \\ \hline
BGA-DGCNN \cite{Uy_2019_ICCV} &0.79 &0.77 &0.58 &0.80 &0.37 &0.76 &\textbf{0.91}  &\textbf{0.76} &0.86 &\textbf{0.91} &\textbf{0.83} &0.76 &0.72 &\textbf{0.80} &0.79 &0.86 &0.78\\\hline
BGA-PN++ \cite{Uy_2019_ICCV} &\textbf{0.80} &0.78 &\textbf{0.64} &\textbf{0.81} &0.46 &0.78 &\textbf{0.91} &0.68 &0.86 &0.89 &0.81 &0.73 &0.81 &0.79 &0.80 &\textbf{0.88} &\textbf{0.86}\\\hline
\textbf{Ours}   &\textbf{0.80} &\textbf{0.79}   &0.61	&0.77	&\textbf{0.52}	&\textbf{0.79}	&0.87	&0.70	&\textbf{0.91}	&\textbf{0.91}	&0.82	&0.75	&\textbf{0.88}	&0.79	&0.78	&0.87	&0.82
\\ \hline
\end{tabular}
\label{tab:scanobjectnn}
}
\end{center}
\end{table*}

\subsubsection{Datasets}
We show the performance of the proposed network on two datasets: a classical ModelNet40~\cite{wu20153d}, which contains synthetic object point clouds, and the recent ScanObjectNN \cite{Uy_2019_ICCV} composed of real-world object point clouds.
\begin{itemize}
 \item \textbf{ModelNet40.}  
As the most widely used benchmark for point cloud analysis, ModelNet40 is popular and authoritative because of its clean shapes, well-constructed dataset, \etc However, it is challenging due to limited and unbalanced training data for various categories. Specifically, the original ModelNet40 consists of 12,311 CAD-generated meshes in 40 categories, of which 9,843 are used for training while the remaining 2,468 are reserved for testing. Moreover, the corresponding data points are uniformly sampled from the mesh surfaces (see Figure~\ref{fig:modelnet}), and then further preprocessed by moving to the origin and scaling into a unit sphere. For our experiments, we only input the 3D coordinates $(x,y,z)$ of 1024 points for each point cloud sample.
 
 \item \textbf{ScanObjectNN.} 
 To further prove the effectiveness and robustness of our classification network, we conduct experiments on ScanObjectNN, a newly published real-world object dataset with about 15k objects in 15 categories. Although it has fewer categories than ModelNet40, it is more practically challenging than its synthetic counterpart due to the complex background, missing parts, and various deformations, as in Figure~\ref{fig:scan}. 
\end{itemize}

\begin{table}
\begin{center}
\caption{Ablation study about different geometric features on \emph{ModelNet40} (\%). (Geometric Point Descriptor: low-level geometric features as in Equation~\ref{equ:nmd}. ABEM (Attentional Back-projection Edge Features Module): high-level geometric features as in Equation~\ref{equ:abem}; prominent: local prominent features only as in Equation~\ref{equ:pfe}; fine-grained: local fine-grained features only as in Equation~\ref{equ:ffe})}
\resizebox{0.9\columnwidth}{!}{
\begin{tabular}{c|c|c|c|c}
\Xhline{3\arrayrulewidth}
\multirow{2}{*}{model} & \multirow{2}{*}{\begin{tabular}[c]{@{}c@{}}Geometric \\ Point Descriptor\end{tabular}} & \multicolumn{2}{c|}{ABEM} & \multirow{2}{*}{overall acc.} \\ \cline{3-4}
                       &                                  & Prominent & Fine-grained &                               \\\hline
0                      & --                            & --       & --           & 92.6                             \\
1                      & --                              & \checkmark       & --           & 93.2                   \\
2                      & --                              & --       & \checkmark           & 92.9                   \\
3                      & --                              & \checkmark       & \checkmark   & 93.4                   \\
4                      & \checkmark                              & --       & --           & 92.7                   \\
\textbf{5}             & \checkmark                      & \checkmark       & \checkmark   & \textbf{93.8}          \\
\Xhline{3\arrayrulewidth}      
\end{tabular}
\label{tab:modules}
}
\end{center}
\end{table}

\subsection{Classification Performance}
\subsubsection{Results on synthetic point clouds}
Table \ref{tab:modelnet} shows the quantitative results on the synthetic ModelNet40 classification benchmark. The result of our network (overall acc: 93.8\% and average class acc: 91.0\%) exceeds state-of-the-art methods comparing under the same given input \ie 1k coordinates only. It is worth mentioning that our approach is even better than some methods using extra input points \eg, DGCNN \cite{wang2019dynamic} with 2k inputs obtained overall acc: 93.5\% and average class acc: 90.7\%. Similarly, our algorithm outperforms SO-Net \cite{li2018so}, which uses rich information such as 5k inputs with normals, achieving an overall accuracy of 93.4\%, and average class accuracy of 90.8\%.  We are also higher than RS-CNN \cite{liu2019relation}, which uses post-processing of a ten votes evaluation arrangement during testing. Rarely do recent methods result in an improvement of more than 0.5\%, while we achieve nearly 1\% improvement compared to the latest work~\cite{Yan_2020_CVPR}. In terms of the network architecture itself, our approach is indicated to be promising and effective for classification.

\subsubsection{Results on real-world point clouds}
For real-world point cloud classification, we use the same network architecture, training strategy, and 1k of 3D coordinates as input. To have fair comparisons with state-of-the-art methods, we conduct the classification experiment with its most challenging variant\footnote{\emph{PB\_T50\_RS}, the hardest case of ScanobjectNN dataset} as in \cite{Uy_2019_ICCV}. We present Table \ref{tab:scanobjectnn} with the accuracies of competing methods on the real-world ScanObjectNN dataset. The results of our network with an overall accuracy of 80.5\% and an average class accuracy of 77.8\% have significantly improved the classification accuracy on the benchmark. We perform better than other methods in 6 out of 15 categories, and for challenging cases like box or display, we improve the accuracy by more than 10\%. Furthermore, our approach performs even better than DGCNN \cite{wang2019dynamic} and PointNet++ \cite{qi2017pointnet++} with background-aware network (BGA) \cite{Uy_2019_ICCV} , which is specially designed for this dataset.

\begin{table}
\begin{center}
\caption{Ablation study about different Geometric Point Descriptor $\widetilde{p}_i$ forms on \emph{ModelNet40} classification accuracy (\%). ($p$: $(x,y,z)$, $n$: $normal$, $e$: $edge$, $l$: $|edge|$.)}
\resizebox{\columnwidth}{!}{
\begin{tabular}{c|lc|c}
\Xhline{3\arrayrulewidth}
model    & Geometric Point Descriptor & length & overall acc. \\\hline
1        & $\widetilde{p}_i = (p_i)$    &3     & 93.4         \\
2        & $\widetilde{p}_i = (p_i, n_i, {l_1}, {l_2})$ & 8 & 93.5         \\
3        & $\widetilde{p}_i = (p_i, e_1, e_2, {l_1}, {l_2})$ &11 & 93.5      \\
4        & $\widetilde{p}_i = (p_i, p_{j1}, p_{j2}, {l_1}, {l_2})$ &11 & 93.3      \\
5        & $\widetilde{p}_i = (p_i, n_i, e_1, e_2)$  &12     & 93.7      \\ 
\textbf{6}        & $\bm{\widetilde{p}_i = (p_i, n_i, e_1, e_2, {l_1}, {l_2})}$ &\textbf{14} & \textbf{93.8}  \\ 
7        & $\widetilde{p}_i = (p_i, n_i, n_{j1}, n_{j2}, e_1, e_2)$  &18     & 93.3      \\
8        & $\widetilde{p}_i = (p_i, n_i, n_{j1}, n_{j2}, e_1, e_2, e_3, l_1, l_2, l_3)$  &24     & 93.1      \\
\Xhline{3\arrayrulewidth}      
\end{tabular}
\label{tab:nmd}
}
\end{center}
\end{table}

Despite the fact that the ScanObjectNN dataset contains hard cases for point cloud classification, the main difficulties are effectively managed by our approach. For example, the remaining background points in real data may confuse the network because they are irrelevant to the shape structure. However, the CAA module is intended to re-weight the points according to their channel affinities, by which the importance of background points can be reduced. Further, in order to form a global representation of a point cloud, we learn both local prominent and fine-grained features using our ABEM module. Such comprehensive local context for a global representation can alleviate the side effects of the missing parts in real point clouds. Although some deformations or distortions may exist in real point cloud data, our network can be relatively robust since the geometric point descriptor is designed to enhance the relations between points and enrich the low-level geometric information. 

To further demonstrate the effectiveness and robustness, we also apply ${F_1}$ score, the harmonic mean of the \emph{precision} and \emph{recall}, as another quantitative measurement comparing with the best two state-of-the-art methods on ScanObjectNN official leaderboard~\cite{scanWeb}: DGCNN~\cite{wang2019dynamic} and PointCNN~\cite{li2018pointcnn}, as well as the BGA-based methods \footnote{pre-trained models in \cite{Uy_2019_ICCV}, downloaded from \url{https://github.com/hkust-vgd/scanobjectnn}} in \cite{Uy_2019_ICCV}. As the bottom rows of Table \ref{tab:scanobjectnn} show, our approach achieves better results in terms of both overall and average class $F_1$ scores. Specifically, we have a higher $F_1$ score on 5 out of 15 testing categories. In general, our network has a better balance between the \emph{precision} and \emph{recall} targeting real-world point cloud classification. As stated before, the point cloud analysis aims to solve practical problems. The excellent performance on real-world point clouds is a strong affirmation of our work.

\begin{table}
\begin{center}
\caption{Classification accuracy (\%) for various \emph{Attention} modules on \emph{ModelNet40}. (\emph{SE} and \emph{Non-local} are adapted from the 2D case by letting the 3D point space be equal to the 2D spatial space. $N$ is the number of points, and $C$ is the number of channels. Commonly in deep learning-based point cloud networks, $N$ is much larger than $C$.)}
\resizebox{\columnwidth}{!}{
\begin{tabular}{c|c|c|c}
\Xhline{3\arrayrulewidth}
module    & operating space & attention map size & overall acc. \\\hline
SE \cite{hu2018squeeze}        & Channel-wise    & $1\times C$     & 93.4         \\
Non-local \cite{wang2018non}        & Point-wise & $N\times N$ & 93.3          \\
L2G-SA \cite{liu2019l2g}        & Point-wise & $N\times N$ & 93.2       \\
Pointwise-SA \cite{xie2018attentional,feng2019point}        & Point-wise  & $N\times N$     & 93.2       \\
PSA \cite{zhao2018psanet} & Point-wise & $2\times(N\times N)$ & 93.3       \\
Criss-Cross \cite{huang2019ccnet} & Point-wise & $N\times N$ & 93.5       \\
\textbf{CAA (ours)}        & Channel-wise & $C\times C$ & \textbf{93.8}  \\\Xhline{3\arrayrulewidth}      
\end{tabular}
\label{tab:attention}
}
\end{center}
\end{table}
\subsection{Ablation Studies}
To verify the functions and effectiveness of different components in our network, we conduct ablation studies about the proposed modules in this work. Besides, we check the network complexity, visualize the learned features, and test preliminary segmentation performance.

\subsubsection{Effects of geometric features}
Table \ref{tab:modules} shows the results of the ablation study concerning the effects of different geometric features learned by our network. It can be observed from model 1,2,3 that the high-level geometric features learned by our Attentional Back-projection Edge Features Module (ABEM) can benefit point cloud classification. Comparing with the fined-grained features in Equation~\ref{equ:ffe}, the local prominent features in Equation~\ref{equ:pfe} contribute more to the high-level geometric features from ABEM. However, it is worth noting that the effect of increasing low-level geometric information alone (\ie, model 4) is minimal because the network lacks modules in high-level space to utilize the provided information fully. In contrast, once we enrich the geometric features from both levels, as in model 5, the performance can significantly boost. In general, our geometric network is tested to work as a whole with inherent relations between low-level and high-level geometric features.

\begin{table}
\begin{center}
\caption{Complexity of classification network on \emph{ModelNet40}. ($^*$the inference time of our model running on GeForce GTX 2080Ti can be reduced to \emph{17.5ms})}
\resizebox{0.9\columnwidth}{!}{
\begin{tabular}{l|ccc}
\Xhline{3\arrayrulewidth}
method    & model size (MB) & time (ms) & overall acc. (\%) \\\hline
PointNet \cite{qi2017pointnet}        & 40 & \textbf{16.6} & 89.2         \\
PointNet++ \cite{qi2017pointnet++}        & \textbf{12} &163.2 & 90.7      \\
PCNN \cite{atzmon2018point}        & 94  &117.0     & 92.3      \\
DGCNN \cite{wang2019dynamic}        & 21 &27.2 & 92.9  \\
\textbf{Ours}       & 34 &$32.2^\star$ & \textbf{93.8}  \\\Xhline{3\arrayrulewidth}      
\end{tabular}
\label{tab:complexity}
}
\end{center}
\end{table}

\subsubsection{Forms of Geometric Point Descriptor}
\label{sec:exp_nmd}
Although estimating low-level geometric features is intuitive, the expanded information can provide more geometric cues for the following processing in high-level space. To avoid possible overfitting caused by involving the redundant information, we investigate various combinations of low-level geometric items to find a better form for the Geometric Point Descriptor. Besides the items introduced in Equation~\ref{equ:nmd}, here we investigate more features:

\begin{small}
$$
 \begin{cases}
    {p_{j1}} = (x_{j1},y_{j1},z_{j1}); &\mbox{$p_{j1}\in\mathbb{R}^{3}$},\\
    {p_{j2}} = (x_{j2},y_{j2},z_{j2}); &\mbox{$p_{j2}\in\mathbb{R}^{3}$},\\
    {edge}_3 = {p_{j1}-p_{j2}}; &\mbox{${edge}_3\in\mathbb{R}^{3}$},\\
    {length_3} = |{edge}_2|; &\mbox{${length_3}\in\mathbb{R}^{1}$},\\
    normal_i = {{edge}_1}\times{{edge}_2}; &\mbox{$normal_i\in\mathbb{R}^{3}$},\\
    normal_{j1} = {(-{edge}_1)}\times{(-{edge}_3)}; &\mbox{$normal_{j1}\in\mathbb{R}^{3}$},\\
    normal_{j2} = {(-{edge}_2)}\times{{edge}_3}; &\mbox{$normal_{j2}\in\mathbb{R}^{3}$}.
 \end{cases}
$$
\end{small}

In general, the geometric point descriptor is explicitly formed in 3D space, containing four types of low-level geometric features: the coordinates, the edges, the normals, and the length of edges. As a fundamental feature, the point's coordinates (\ie, position) can intuitively indicate the global information. On the other hand, the point's edges can imply more local information by involving its neighbors' relative positions, and the lengths of edges can partly estimate the density distribution of the point's neighborhood. In addition, we calculate the point's normal using the cross-product of its edge vectors to enhance the descriptor's robustness against possible deformations such as scaling, translation, rotation, \etc However, it is not always optimal to combine all of these features since it may incorporate redundant information: for example, model~7 in Table~\ref{tab:nmd} contains the normal vectors of neighbors, \ie, $n_{j1}$ and $n_{j2}$, which do not help describe the low-level geometry of the point $p_i$; and for model~8, the edge clues between two neighbors (\ie, $e_3$ and $l_3$) share similar information to $e_1$, $l_1$ and $e_2$, $l_2$.

According to the experiments shown in Table~\ref{tab:nmd}, we conclude the best form is model 6, as in Equation \ref{equ:nmd}, since it can adequately represent the low-level geometric features of the estimated triangle face while in a relatively compact form.

\begin{figure*}
\begin{center}
\subfigure[Examples of the features learned by ABEMs in different layers of our network. Shallow layers focus on edges/corners, while deep layers cover semantically meaningful (discriminative) parts that can help to identify the class label of a certain point cloud: \eg, the tail/wings of a plane (1st row in the figure), the fingerboard of a guitar (2nd row in the figure), or the head/arms/legs of a human (3rd row in the figure), \etc Further, prominent and fine-grained feature encoding branches capture complementary high-level geometric features, which are crucial for comprehensive point clouds representations.]{                    
    \includegraphics[width=0.85\textwidth]{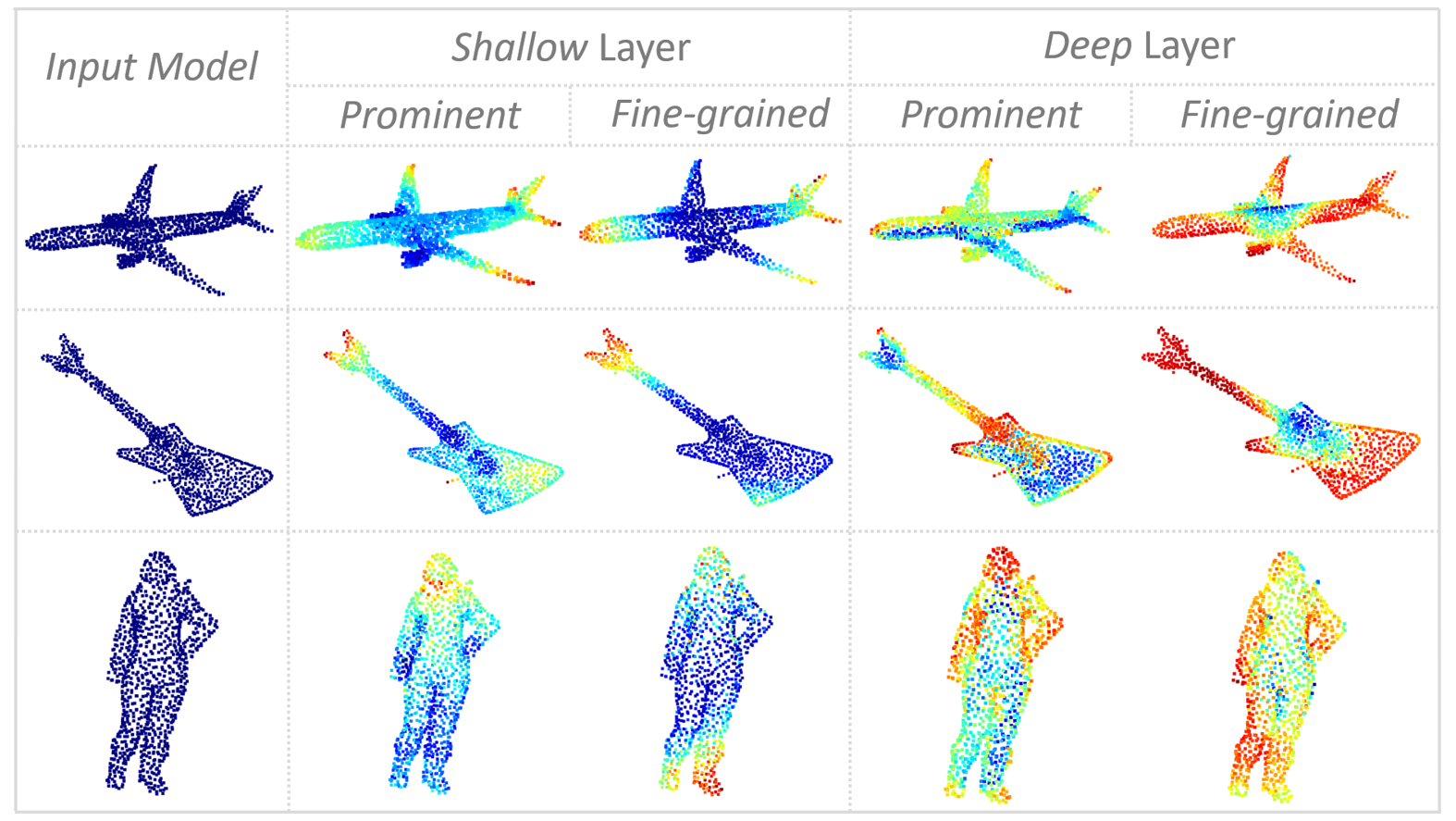}\label{fig:abem_vis}               
    }
\subfigure[Changes to the feature map processed by the Channel-wise Affinity Attention (CAA) module. The CAA module can enhance distinct channels and avoid redundancy by distributing weights along channels according to the calculated \emph{Channel Affinity} in Equation~\ref{equ:affinity}. After CAA processing, the feature map emphasizes the semantically meaningful parts such as the headboard/mattress of a bed (1st column in the figure), the back of a sofa (2nd column in the figure), the hood/chassis of a car (3rd column in the figure), or the keyboard of a piano (4th column in the figure), while the responses of simple edges/corners are reduced.]{                    
    \includegraphics[width=0.9\textwidth]{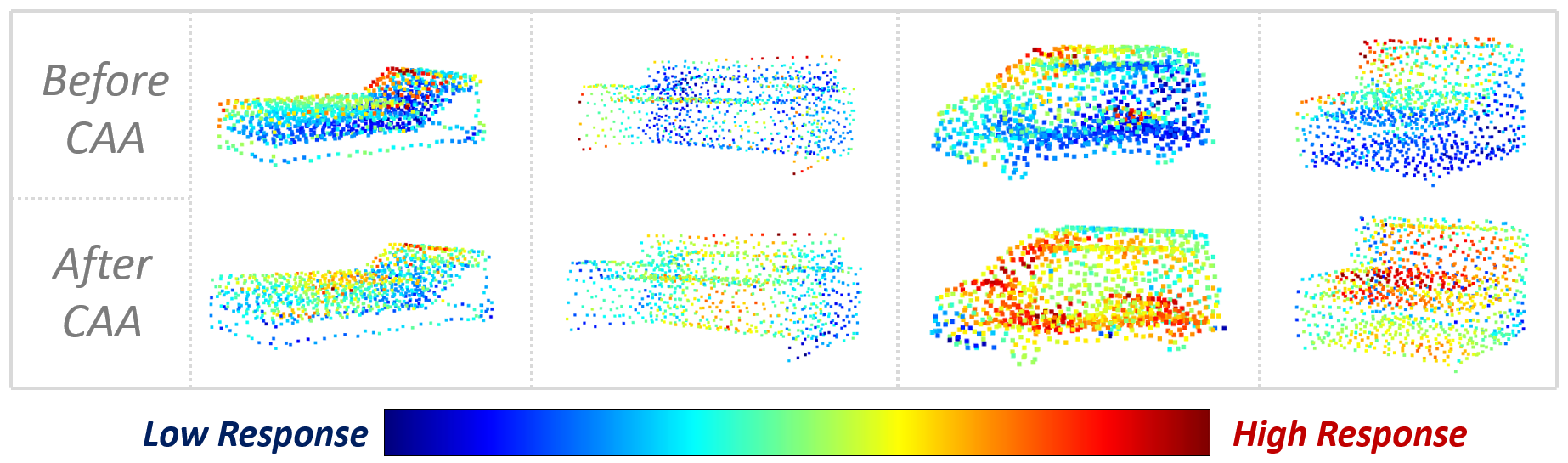}\label{fig:caa_vis}               
    }
\end{center}
\caption{Visualizations of the intermediate feature maps. Figure~\ref{fig:abem_vis} shows the features learned by ABEMs, and Figure~\ref{fig:caa_vis} depicts the CAA effects.}
\label{fig:vis}
\end{figure*}

However, the effects of the current geometric point descriptor may be marginal in some downstream tasks (\eg, semantic segmentation, detection) for large-scale data since the point distributions can be quite uneven as the data is scanned from real scenes. Even for similar underlying surfaces, possible distortions can cause inconsistent geometric relations. As a consequence, the representation ability of our low-level geometry-based descriptor will be affected. Hence, we intend to further investigate a generalized form of the geometric point descriptor in the future.

\subsubsection{Attention modules for point cloud classification}
To have fair comparisons, we replace all CAA modules in our network with other advanced attention modules. Besides the accuracies, we also extract the size of the intermediate attention map to estimate the corresponding memory requirement of each approach. As Table \ref{tab:attention} presents, comparing with classical 2D adapted and state-of-the-art 3D designed attention modules, our Channel Affinity Attention module shows both effectiveness and efficiency.

\begin{figure*}
\begin{center}
\includegraphics[width=0.85\textwidth]{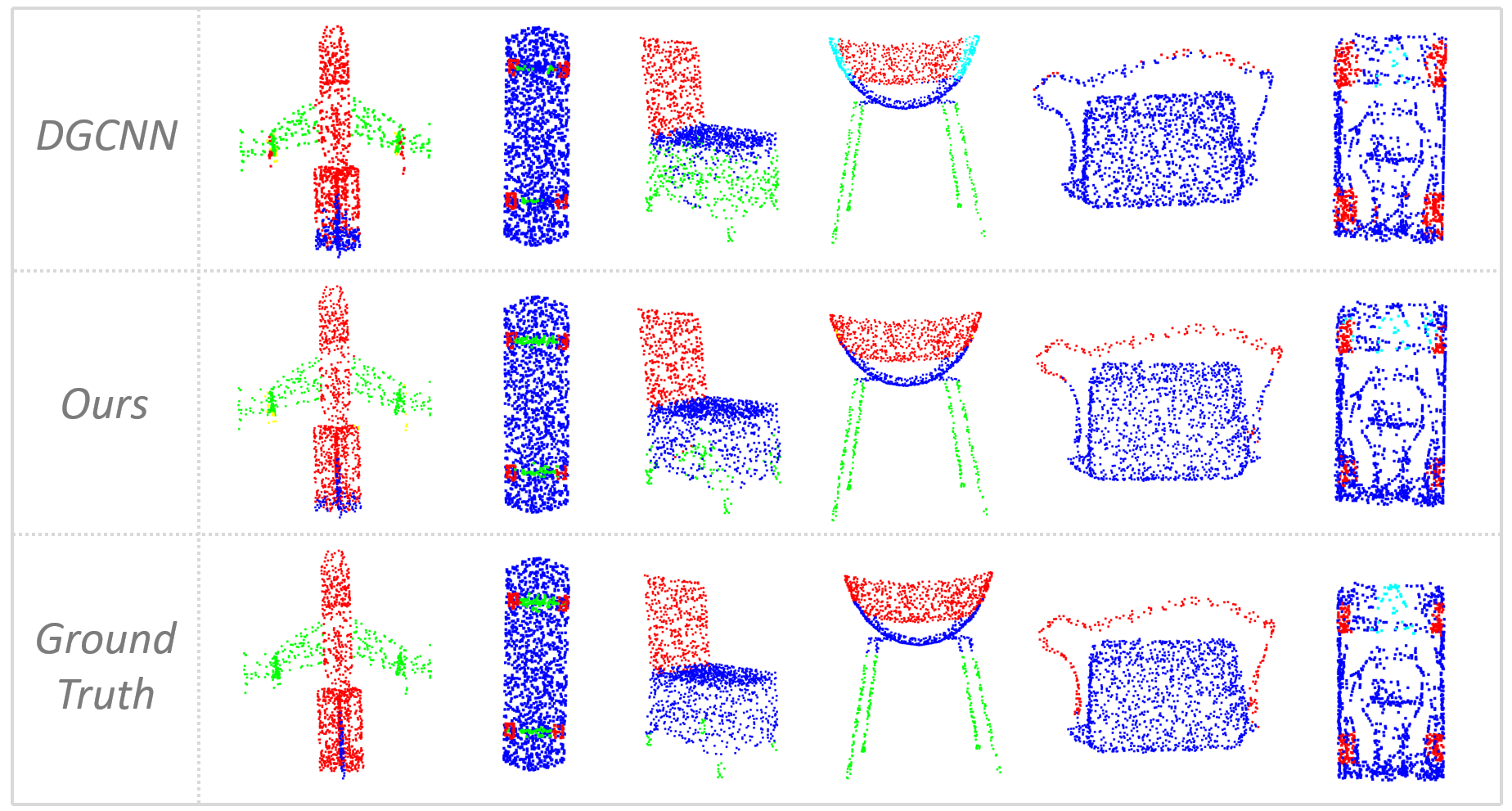}
\end{center}
  \caption{Examples of the point cloud segmentation results on \emph{ShapeNet Part}. The first row shows the results of DGCNN~\cite{wang2019dynamic}; the second row exhibits the results of our approach, while the ground truths are in the last row.}
\label{fig:seg_vis}
\end{figure*}

\begin{table*}
\begin{center}
\caption{Part segmentation results (mIoU(\%)) on \emph{ShapeNet Part} dataset. }
\resizebox{\textwidth}{!}{
\begin{tabular}{||c| c c c c c c c c c c c c c c c c |c||}
\hline
 & air & bag & cap  & car  & chair  & ear & guitar & knife & lamp & laptop & moto & mug & pistol  & rocket & skate & table &\textbf{overall}  \\
 & plane &  &   &   &   & phone &  &  &  &  & bike &  &   &  & board &  &\textbf{mIoU} \\\hline\hline
\# shapes  & 2690   & 76 & 55  & 898  & 3758  & 69    & 787  & 392  & 1547     & 451  & 202  & 184   & 283  & 66    & 152  & 5271 & 16881    \\ \hline\hline
PointNet \cite{qi2017pointnet} &83.4 &78.7 &82.5 &74.9 &89.6 &73.0 &91.5 &85.9 &80.8 &95.3 &65.2 &93.0 &81.2 &57.9 &72.8 &80.6 & 83.7    \\ \hline
PointNet++ \cite{qi2017pointnet++}  & 82.4  & 79.0  & 87.7  & 77.3  & 90.8  & 71.8  & 91.0  & 85.9  & 83.7  & 95.3  & 71.6  & 94.1  & 81.3  & 58.7  & \textbf{76.4}  & 82.6 & 85.1    \\ \hline
A-SCN \cite{xie2018attentional} &83.8 &80.8 &83.5 &79.3 &90.5 &69.8 &\textbf{91.7} &86.5 &82.9 &\textbf{96.0} &69.2 &93.8 &82.5 &62.9 &74.4 &80.8 & 84.6    \\ \hline
SpiderCNN \cite{xu2018spidercnn} &83.5 &81.0 &87.2 &77.5 &90.7 &\textbf{76.8} &91.1 &87.3 &83.3 &95.8 &70.2 &93.5 &82.7 &59.7 &75.8 &82.8 &85.3\\ \hline
SO-Net \cite{li2018so} &81.9 &83.5 &84.8 &78.1 &90.8 &72.2 &90.1 &83.6 &82.3 &95.2 &69.3 &94.2 &80.0 &51.6 &72.1 &82.6 &84.6\\ \hline
PCNN \cite{atzmon2018point} &82.4 &80.1 &85.5 &79.5 &90.8 &73.2 &91.3 &86.0 &85.0 &95.7 &73.2 &94.8 &83.3 &51.0 &75.0 &81.8 &85.1\\ \hline
DGCNN \cite{wang2019dynamic}  & 84.0   & 83.4 & 86.7  & 77.8  & 90.6  & 74.7    & 91.2  & 87.5  & 82.8     & 95.7  & 66.3  & 94.9   & 81.1  & \textbf{63.5}    & 74.5  & 82.6 & 85.2    \\ \hline
RGCNN \cite{te2018rgcnn}  & 80.2   & 82.8 & \textbf{92.6}  & 75.3  & 89.2  & 73.7    & 91.3  & 88.4  & 83.3     & \textbf{96.0}  & 63.9  & \textbf{95.7}   & 60.9  & 44.6    & 72.9  & 80.4 & 84.3    \\ \hline
PointASNL \cite{Yan_2020_CVPR} &84.1 &\textbf{84.7} &87.9 &\textbf{79.7} &\textbf{92.2} &73.7 &91.0 &87.2 &84.2 &95.8 &\textbf{74.4} &95.2 &81.0 &63.0 &76.3 &83.2 &\textbf{86.1}\\
\hline\hline
\textbf{Ours}  &\textbf{84.5} &82.2 &86.8 &78.9 &91.1 &74.5 &91.4 &\textbf{89.0} &\textbf{84.5} &95.5 &69.6 &94.2 &\textbf{83.4} &57.8 &75.5 &\textbf{83.5} &85.9   \\ \hline
\end{tabular}
\label{tab:seg}
}
\end{center}
\end{table*}

\subsubsection{Network Complexity}
Since our network consists of several key modules, it is necessary to compare our model's complexity with other state-of-the-art methods. Specifically, we apply similar measurements as in \cite{wang2019dynamic}: model size and inference time. To be fair with the competing methods, we test under the same conditions (GeForce GTX 1080, batch size = 1) as in \cite{wang2019dynamic}. Although we apply complex operations like FC layers or $knn$ searching as others, we manage to simplify the complexity by sharing weights, removing alignment blocks~\cite{qi2017pointnet,wang2019dynamic}, compacting features, \etc Table \ref{tab:complexity} indicates that we have a relatively good compromise between accuracy and efficiency. Furthermore, our model's inference time running on GeForce RTX 2080Ti is about \emph{17.5ms}, and we expect to optimize it for real-time applications further.

\subsubsection{Visualization}
From Figure \ref{fig:abem_vis} we can visualize corresponding learned features by the prominent feature encoding branch (\ie, Equation~\ref{equ:pfe}) and the fine-grained feature encoding branch (\ie, Equation~\ref{equ:ffe}) of our Attentional Back-projection Edge Features Module (\ie, ABEM in Equation~\ref{equ:abem}) respectively in different layers of the network. Particularly, all examples show the property of CNN: the shallow layers have a higher impact on simpler features \eg, edges, corners, \etc while deep layers connect those simpler features for more semantically specific parts. As we stated before, prominent feature encoding branch extracts main local context while fine-grained feature encoding branch helps to capture missing details. From the figure, we can observe that fine-grained features complement prominent features as expected.

As stated in Section~\ref{ref:CAA}, the proposed Channel-wise Affinity Attention (CAA) module aims to refine the feature map along channels. From Figure \ref{fig:caa_vis}, we observe the effects of CAA on the feature map. The CAA module enhances the distinct channels based on calculated \emph{ Channel Affinity} (\ie, $\mathcal{A}_{{d}\times{d}}$ in Equation~\ref{equ:affinity}), which contributes to more attention on semantically meaningful parts. On the other hand, the responses of simple features like edges/corners, are effectively weakened. In general, CAA refines the feature map for more compact and effective representations.

\subsubsection{Part Segmentation}
We further demonstrate the advantages of our approach on the point cloud segmentation task. Specifically, we adopt the ShapeNet Part Segmentation dataset~\cite{yi2016scalable}, which includes 16,881 instance-level point clouds of 16 different classes, with each point being labeled as one of the 50 candidate parts. In terms of the training and test sets, we follow the official data split in~\cite{chang2015shapenet}. 

\begin{table*}
\begin{center}
\caption{3D object detection results (\%) on \emph{SUN RGB-D V2}~\cite{song2015sun} dataset. We adopt the Hough voting structure from VoteNet~\cite{qi2019deep}, while replacing the backbone with different point cloud networks. The performances are evaluated using the metric of the average precision (mAP) with 3D IoU threshold 0.25~\cite{qi2019deep, song2015sun}.}
\resizebox{0.95\textwidth}{!}{
\begin{tabular}{||c| c c c c c c c c c c |c||}
\hline
backbone &bed &table &sofa &chair &toilet &desk &dresser &night-stand &book-shelf &bathtub &\textbf{mAP} \\ \hline\hline
PointNet \cite{qi2017pointnet} &46.2&23.9&28.3&52.1&38.4&9.5&1.4&4.6&3.6&17.8&22.6    \\ \hline
PointNet++ \cite{qi2017pointnet++}  &84.8&\textbf{51.3}&\textbf{68.5}&\textbf{77.0}&88.8&\textbf{25.3}&27.3&60.6&32.2&72.2&58.8    \\ \hline
DGCNN \cite{wang2019dynamic}  &84.6&46.0&60.0&76.8&\textbf{89.1}&19.5&\textbf{31.2}&54.8&24.6&72.8&55.9    \\ \hline
RS-CNN \cite{liu2019relation}  &\textbf{85.7}&47.3&64.9&\textbf{77.0}&85.2&24.9&27.9&60.5&30.4&77.4&58.1    \\ \hline\hline
\textbf{Ours}   &84.8&51.0&64.1&\textbf{77.0}&88.1&24.4&\textbf{31.2}&\textbf{62.9}&\textbf{32.4}&\textbf{77.5}&\textbf{59.3}   \\ \hline
\end{tabular}
\label{tab:detection}
}
\end{center}
\end{table*}

Generally, our architecture for segmentation is in a basic fully convolutional network~\cite{long2015fully} form. By using the backbone of our classification network, we concatenate the learned point features from different scales. After attaching the max-pooled global feature and one-hot-labels of objects, we apply fully connected layers to regress the confidence scores for 50 possible part classes of each point. Particularly, we only input the 3D coordinates of 2048 points for each point cloud for training and evaluate the performance based on the Intersection-over-Union (\ie, IoU) metric. The IoU of the shape is calculated by the mean value of IoUs of all parts in that shape, while the Mean Intersection over Union (\ie, mIoU) is the average of IoUs for all testing shapes.

Table \ref{tab:seg} presents the experimental results of some state-of-the-art methods. In terms of instance mIoU, we outperform most of the competitors while we are slightly behind the latest PointASNL~\cite{Yan_2020_CVPR}, which forms an upsampling and downsampling architecture targeting point cloud segmentation tasks. Regarding the mIoU of each class, our basic fully convolutional network for segmentation won 5 out of a total 16 tested classes. Moreover, we can easily find in Figure~\ref{fig:seg_vis} that our approach performs better than DGCNN~\cite{wang2019dynamic} on the ShapeNet Part Segmentation dataset.

\subsubsection{Object Detection}
In addition, we conduct 3D object detection experiments to validate our network's generalization ability for more downstream tasks. Based on the Hough voting module proposed by VoteNet~~\cite{qi2019deep}, we apply different state-of-the-art models as the network backbone running on SUN RGB-D V2~\cite{song2015sun} datatset.

Specifically, the SUN RGB-D V2 dataset has 5285 point clouds in the training set and other 5050 samples as the testing set, where the coordinates of points are used as the input. Following the same metric used in~\cite{qi2019deep, song2015sun}, we evaluate the performance by calculating the mean average precision (mAP) with an IoU threshold of 0.25. Table~\ref{tab:detection} presents the detailed results of different backbone networks. Particularly, our network achieves a higher overall score of 59.3\% mAP@0.25 and performs better than others in five out of ten testing categories. In general, our network shows excellent potential in 3D object detection.

% \begin{table}
% \begin{center}
% \caption{3D object detection results (\%) on \emph{SUN RGB-D V2} dataset. We adopt the Hough voting structure of VoteNet~\cite{qi2019deep} for detection, while replacing its backbone with different point cloud analysis networks. The average precision (mAP) with 3D IoU threshold 0.25 and 0.5 are reported respectively as in~\cite{song2015sun}.}
% \resizebox{0.7\columnwidth}{!}{
% \begin{tabular}{l|cc}
% \Xhline{3\arrayrulewidth}
% backbone    & mAP@0.25 & mAP@0.5 \\\hline
% PointNet \cite{qi2017pointnet}        &22.6 &5.1         \\
% PointNet++ \cite{qi2017pointnet++}  &58.8 &\textbf{35.1}      \\
% DGCNN \cite{wang2019dynamic}        &55.9 &29.4  \\
% RS-CNN \cite{liu2019relation}        &58.1 &33.3  \\
% \textbf{Ours}       &\textbf{59.3} &34.2  \\\Xhline{3\arrayrulewidth}      
% \end{tabular}
% %\label{tab:detection}
% }
% \end{center}
% \end{table}

\section{Conclusion}
\label{sec:concl}
In this paper, we propose a new point-based CNN network targeting point cloud classification. To leverage more geometric features from scattered point clouds, we explicitly apply the Geometric Point Descriptor to estimate geometric clues in low-level space. Accordingly, the Attentional Back-projection Edge Features Module can automatically assist learning a better point cloud representation using local geometric context in high-level space. Besides, the Channel-wise Affinity Attention module further refines the learned feature map by focusing on distinct channels. To compare our method with other state-of-the-art networks, we conduct experiments on synthetic and real-world point cloud datasets. Also, we validate the properties of the proposed modules through necessary ablation studies and visualizations. The results show the effectiveness and robustness of our approach. More tasks, \eg scene parsing, instance segmentation, will be explored in the future.

% or


% \appendices

% % you can choose not to have a title for an appendix
% % if you want by leaving the argument blank
% \section{}

% % use section* for acknowledgment
% \section*{Acknowledgment}


\ifCLASSOPTIONcaptionsoff
  \newpage
\fi



%\bibliographystyle{IEEEtran}
\bibliographystyle{IEEEtran}
\bibliography{egbib}

\begin{thebibliography}{10}
\providecommand{\url}[1]{#1}
\csname url@samestyle\endcsname
\providecommand{\newblock}{\relax}
\providecommand{\bibinfo}[2]{#2}
\providecommand{\BIBentrySTDinterwordspacing}{\spaceskip=0pt\relax}
\providecommand{\BIBentryALTinterwordstretchfactor}{4}
\providecommand{\BIBentryALTinterwordspacing}{\spaceskip=\fontdimen2\font plus
\BIBentryALTinterwordstretchfactor\fontdimen3\font minus
  \fontdimen4\font\relax}
\providecommand{\BIBforeignlanguage}[2]{{%
\expandafter\ifx\csname l@#1\endcsname\relax
\typeout{** WARNING: IEEEtran.bst: No hyphenation pattern has been}%
\typeout{** loaded for the language `#1'. Using the pattern for}%
\typeout{** the default language instead.}%
\else
\language=\csname l@#1\endcsname
\fi
#2}}
\providecommand{\BIBdecl}{\relax}
\BIBdecl

\bibitem{chen2013point}
J.-Y. Chen, C.-H. Lin, P.-C. Hsu, and C.-H. Chen, ``Point cloud encoding for 3d
  building model retrieval,'' \emph{IEEE transactions on multimedia}, vol.~16,
  no.~2, pp. 337--345, 2013.

\bibitem{wu2016fast}
P.~Wu, Y.~Liu, M.~Ye, J.~Li, and S.~Du, ``Fast and adaptive 3d reconstruction
  with extensively high completeness,'' \emph{IEEE Transactions on Multimedia},
  vol.~19, no.~2, pp. 266--278, 2016.

\bibitem{de2018graph}
P.~de~Oliveira~Rente, C.~Brites, J.~Ascenso, and F.~Pereira, ``Graph-based
  static 3d point clouds geometry coding,'' \emph{IEEE Transactions on
  Multimedia}, vol.~21, no.~2, pp. 284--299, 2018.

\bibitem{valsesia2020learning}
D.~Valsesia, G.~Fracastoro, and E.~Magli, ``Learning localized representations
  of point clouds with graph-convolutional generative adversarial networks,''
  \emph{IEEE Transactions on Multimedia}, 2020.

\bibitem{zhang2020pointhop}
M.~Zhang, H.~You, P.~Kadam, S.~Liu, and C.-C.~J. Kuo, ``Pointhop: An
  explainable machine learning method for point cloud classification,''
  \emph{IEEE Transactions on Multimedia}, 2020.

\bibitem{blais2004review}
F.~Blais \emph{et~al.}, ``Review of 20 years of range sensor development,''
  \emph{Journal of electronic imaging}, vol.~13, no.~1, pp. 231--243, 2004.

\bibitem{jaboyedoff2012use}
M.~Jaboyedoff, T.~Oppikofer, A.~Abell{\'a}n, M.-H. Derron, A.~Loye, R.~Metzger,
  and A.~Pedrazzini, ``Use of lidar in landslide investigations: a review,''
  \emph{Natural hazards}, vol.~61, no.~1, pp. 5--28, 2012.

\bibitem{deng2009imagenet}
J.~Deng, W.~Dong, R.~Socher, L.-J. Li, K.~Li, and L.~Fei-Fei, ``Imagenet: A
  large-scale hierarchical image database,'' in \emph{2009 IEEE conference on
  computer vision and pattern recognition}.\hskip 1em plus 0.5em minus
  0.4em\relax Ieee, 2009, pp. 248--255.

\bibitem{yang2018foldingnet}
Y.~Yang, C.~Feng, Y.~Shen, and D.~Tian, ``Foldingnet: Point cloud auto-encoder
  via deep grid deformation,'' in \emph{Proceedings of the IEEE Conference on
  Computer Vision and Pattern Recognition}, 2018, pp. 206--215.

\bibitem{roveri2018network}
R.~Roveri, L.~Rahmann, C.~Oztireli, and M.~Gross, ``A network architecture for
  point cloud classification via automatic depth images generation,'' in
  \emph{Proceedings of the IEEE Conference on Computer Vision and Pattern
  Recognition}, 2018, pp. 4176--4184.

\bibitem{chen2018sampled}
J.~Chen, Y.~K. Cho, and J.~Ueda, ``Sampled-point network for classification of
  deformed building element point clouds,'' in \emph{2018 IEEE International
  Conference on Robotics and Automation (ICRA)}.\hskip 1em plus 0.5em minus
  0.4em\relax IEEE, 2018, pp. 2164--2169.

\bibitem{Uy_2019_ICCV}
M.~A. Uy, Q.-H. Pham, B.-S. Hua, T.~Nguyen, and S.-K. Yeung, ``Revisiting point
  cloud classification: A new benchmark dataset and classification model on
  real-world data,'' in \emph{The IEEE International Conference on Computer
  Vision (ICCV)}, October 2019.

\bibitem{Li_2020_CVPR}
R.~Li, X.~Li, P.-A. Heng, and C.-W. Fu, ``Pointaugment: An auto-augmentation
  framework for point cloud classification,'' in \emph{IEEE/CVF Conference on
  Computer Vision and Pattern Recognition (CVPR)}, June 2020.

\bibitem{Nezhadarya_2020_CVPR}
E.~Nezhadarya, E.~Taghavi, R.~Razani, B.~Liu, and J.~Luo, ``Adaptive
  hierarchical down-sampling for point cloud classification,'' in
  \emph{IEEE/CVF Conference on Computer Vision and Pattern Recognition (CVPR)},
  June 2020.

\bibitem{schnabel2007efficient}
R.~Schnabel, R.~Wahl, and R.~Klein, ``Efficient ransac for point-cloud shape
  detection,'' in \emph{Computer graphics forum}, vol.~26, no.~2.\hskip 1em
  plus 0.5em minus 0.4em\relax Wiley Online Library, 2007, pp. 214--226.

\bibitem{mitra2004registration}
N.~J. Mitra, N.~Gelfand, H.~Pottmann, and L.~Guibas, ``Registration of point
  cloud data from a geometric optimization perspective,'' in \emph{Proceedings
  of the 2004 Eurographics/ACM SIGGRAPH symposium on Geometry
  processing}.\hskip 1em plus 0.5em minus 0.4em\relax ACM, 2004, pp. 22--31.

\bibitem{rusu2009fast}
R.~B. Rusu, N.~Blodow, and M.~Beetz, ``Fast point feature histograms (fpfh) for
  3d registration,'' in \emph{2009 IEEE International Conference on Robotics
  and Automation}.\hskip 1em plus 0.5em minus 0.4em\relax IEEE, 2009, pp.
  3212--3217.

\bibitem{vosselman20013d}
G.~Vosselman, S.~Dijkman \emph{et~al.}, ``3d building model reconstruction from
  point clouds and ground plans,'' \emph{International archives of
  photogrammetry remote sensing and spatial information sciences}, vol.~34, no.
  3/W4, pp. 37--44, 2001.

\bibitem{guo2019deep}
Y.~Guo, H.~Wang, Q.~Hu, H.~Liu, L.~Liu, and M.~Bennamoun, ``Deep learning for
  3d point clouds: A survey,'' 2019.

\bibitem{su2015multi}
H.~Su, S.~Maji, E.~Kalogerakis, and E.~Learned-Miller, ``Multi-view
  convolutional neural networks for 3d shape recognition,'' in
  \emph{Proceedings of the IEEE international conference on computer vision},
  2015, pp. 945--953.

\bibitem{maturana2015voxnet}
D.~Maturana and S.~Scherer, ``Voxnet: A 3d convolutional neural network for
  real-time object recognition,'' in \emph{2015 IEEE/RSJ International
  Conference on Intelligent Robots and Systems (IROS)}.\hskip 1em plus 0.5em
  minus 0.4em\relax IEEE, 2015, pp. 922--928.

\bibitem{qi2017pointnet}
C.~R. Qi, H.~Su, K.~Mo, and L.~J. Guibas, ``Pointnet: Deep learning on point
  sets for 3d classification and segmentation,'' in \emph{Proceedings of the
  IEEE Conference on Computer Vision and Pattern Recognition}, 2017, pp.
  652--660.

\bibitem{liu2019relation}
Y.~Liu, B.~Fan, S.~Xiang, and C.~Pan, ``Relation-shape convolutional neural
  network for point cloud analysis,'' in \emph{Proceedings of the IEEE
  Conference on Computer Vision and Pattern Recognition}, 2019, pp. 8895--8904.

\bibitem{wang2017cnn}
P.-S. Wang, Y.~Liu, Y.-X. Guo, C.-Y. Sun, and X.~Tong, ``O-cnn: Octree-based
  convolutional neural networks for 3d shape analysis,'' \emph{ACM Transactions
  on Graphics (TOG)}, vol.~36, no.~4, p.~72, 2017.

\bibitem{xu2018spidercnn}
Y.~Xu, T.~Fan, M.~Xu, L.~Zeng, and Y.~Qiao, ``Spidercnn: Deep learning on point
  sets with parameterized convolutional filters,'' in \emph{Proceedings of the
  European Conference on Computer Vision (ECCV)}, 2018, pp. 87--102.

\bibitem{li2018so}
J.~Li, B.~M. Chen, and G.~Hee~Lee, ``So-net: Self-organizing network for point
  cloud analysis,'' in \emph{Proceedings of the IEEE conference on computer
  vision and pattern recognition}, 2018, pp. 9397--9406.

\bibitem{Hu_2020_CVPR}
Q.~Hu, B.~Yang, L.~Xie, S.~Rosa, Y.~Guo, Z.~Wang, N.~Trigoni, and A.~Markham,
  ``Randla-net: Efficient semantic segmentation of large-scale point clouds,''
  in \emph{IEEE/CVF Conference on Computer Vision and Pattern Recognition
  (CVPR)}, June 2020.

\bibitem{weiss1987dynamic}
L.~Weiss, A.~Sanderson, and C.~Neuman, ``Dynamic sensor-based control of robots
  with visual feedback,'' \emph{IEEE Journal on Robotics and Automation},
  vol.~3, no.~5, pp. 404--417, 1987.

\bibitem{kitano2002computational}
H.~Kitano, ``Computational systems biology,'' \emph{Nature}, vol. 420, no.
  6912, pp. 206--210, 2002.

\bibitem{carreira2016human}
J.~Carreira, P.~Agrawal, K.~Fragkiadaki, and J.~Malik, ``Human pose estimation
  with iterative error feedback,'' in \emph{Proceedings of the IEEE conference
  on computer vision and pattern recognition}, 2016, pp. 4733--4742.

\bibitem{haris2018deep}
M.~Haris, G.~Shakhnarovich, and N.~Ukita, ``Deep back-projection networks for
  super-resolution,'' in \emph{Proceedings of the IEEE conference on computer
  vision and pattern recognition}, 2018, pp. 1664--1673.

\bibitem{liu2019hierarchical}
Z.-S. Liu, L.-W. Wang, C.-T. Li, and W.-C. Siu, ``Hierarchical back projection
  network for image super-resolution,'' in \emph{Proceedings of the IEEE
  Conference on Computer Vision and Pattern Recognition Workshops}, 2019, pp.
  0--0.

\bibitem{Li_2019_ICCV}
R.~Li, X.~Li, C.-W. Fu, D.~Cohen-Or, and P.-A. Heng, ``Pu-gan: A point cloud
  upsampling adversarial network,'' in \emph{The IEEE International Conference
  on Computer Vision (ICCV)}, October 2019.

\bibitem{vaswani2017attention}
A.~Vaswani, N.~Shazeer, N.~Parmar, J.~Uszkoreit, L.~Jones, A.~N. Gomez,
  {\L}.~Kaiser, and I.~Polosukhin, ``Attention is all you need,'' in
  \emph{Advances in neural information processing systems}, 2017, pp.
  5998--6008.

\bibitem{wang2018non}
X.~Wang, R.~Girshick, A.~Gupta, and K.~He, ``Non-local neural networks,'' in
  \emph{Proceedings of the IEEE Conference on Computer Vision and Pattern
  Recognition}, 2018, pp. 7794--7803.

\bibitem{hu2018squeeze}
J.~Hu, L.~Shen, and G.~Sun, ``Squeeze-and-excitation networks,'' in
  \emph{Proceedings of the IEEE conference on computer vision and pattern
  recognition}, 2018, pp. 7132--7141.

\bibitem{fu2019dual}
J.~Fu, J.~Liu, H.~Tian, Y.~Li, Y.~Bao, Z.~Fang, and H.~Lu, ``Dual attention
  network for scene segmentation,'' in \emph{Proceedings of the IEEE Conference
  on Computer Vision and Pattern Recognition}, 2019, pp. 3146--3154.

\bibitem{Fang_2019_ICCV}
P.~Fang, J.~Zhou, S.~K. Roy, L.~Petersson, and M.~Harandi, ``Bilinear attention
  networks for person retrieval,'' in \emph{Proceedings of the IEEE/CVF
  International Conference on Computer Vision (ICCV)}, October 2019, pp.
  8030--8039.

\bibitem{Anwar_2019_ICCV}
S.~Anwar and N.~Barnes, ``Real image denoising with feature attention,'' in
  \emph{The IEEE International Conference on Computer Vision (ICCV)}, October
  2019.

\bibitem{xie2018attentional}
S.~Xie, S.~Liu, Z.~Chen, and Z.~Tu, ``Attentional shapecontextnet for point
  cloud recognition,'' in \emph{Proceedings of the IEEE Conference on Computer
  Vision and Pattern Recognition}, 2018, pp. 4606--4615.

\bibitem{feng2019point}
M.~Feng, L.~Zhang, X.~Lin, S.~Z. Gilani, and A.~Mian, ``Point attention network
  for semantic segmentation of 3d point clouds,'' \emph{arXiv preprint
  arXiv:1909.12663}, 2019.

\bibitem{liu2019l2g}
X.~Liu, Z.~Han, X.~Wen, Y.-S. Liu, and M.~Zwicker, ``L2g auto-encoder:
  Understanding point clouds by local-to-global reconstruction with
  hierarchical self-attention,'' in \emph{Proceedings of the 27th ACM
  International Conference on Multimedia}, 2019, pp. 989--997.

\bibitem{mitra2003estimating}
N.~J. Mitra and A.~Nguyen, ``Estimating surface normals in noisy point cloud
  data,'' in \emph{Proceedings of the nineteenth annual symposium on
  Computational geometry}.\hskip 1em plus 0.5em minus 0.4em\relax ACM, 2003,
  pp. 322--328.

\bibitem{merigot2010voronoi}
Q.~M{\'e}rigot, M.~Ovsjanikov, and L.~J. Guibas, ``Voronoi-based curvature and
  feature estimation from point clouds,'' \emph{IEEE Transactions on
  Visualization and Computer Graphics}, vol.~17, no.~6, pp. 743--756, 2010.

\bibitem{kortgen20033d}
M.~K{\"o}rtgen, G.-J. Park, M.~Novotni, and R.~Klein, ``3d shape matching with
  3d shape contexts,'' in \emph{The 7th central European seminar on computer
  graphics}, vol.~3.\hskip 1em plus 0.5em minus 0.4em\relax Budmerice, 2003,
  pp. 5--17.

\bibitem{Yan_2020_CVPR}
X.~Yan, C.~Zheng, Z.~Li, S.~Wang, and S.~Cui, ``Pointasnl: Robust point clouds
  processing using nonlocal neural networks with adaptive sampling,'' in
  \emph{IEEE/CVF Conference on Computer Vision and Pattern Recognition (CVPR)},
  June 2020.

\bibitem{qi2017pointnet++}
C.~R. Qi, L.~Yi, H.~Su, and L.~J. Guibas, ``Pointnet++: Deep hierarchical
  feature learning on point sets in a metric space,'' in \emph{Advances in
  neural information processing systems}, 2017, pp. 5099--5108.

\bibitem{li2018pointcnn}
Y.~Li, R.~Bu, M.~Sun, W.~Wu, X.~Di, and B.~Chen, ``Pointcnn: Convolution on
  x-transformed points,'' in \emph{Advances in Neural Information Processing
  Systems}, 2018, pp. 820--830.

\bibitem{wang2019dynamic}
Y.~Wang, Y.~Sun, Z.~Liu, S.~E. Sarma, M.~M. Bronstein, and J.~M. Solomon,
  ``Dynamic graph cnn for learning on point clouds,'' \emph{ACM Transactions on
  Graphics (TOG)}, vol.~38, no.~5, p. 146, 2019.

\bibitem{engelmann2019dilated}
F.~Engelmann, T.~Kontogianni, and B.~Leibe, ``Dilated point convolutions: On
  the receptive field size of point convolutions on 3d point clouds,'' 2019.

\bibitem{russakovsky2015imagenet}
O.~Russakovsky, J.~Deng, H.~Su, J.~Krause, S.~Satheesh, S.~Ma, Z.~Huang,
  A.~Karpathy, A.~Khosla, M.~Bernstein \emph{et~al.}, ``Imagenet large scale
  visual recognition challenge,'' \emph{International journal of computer
  vision}, vol. 115, no.~3, pp. 211--252, 2015.

\bibitem{woo2018cbam}
S.~Woo, J.~Park, J.-Y. Lee, and I.~So~Kweon, ``Cbam: Convolutional block
  attention module,'' in \emph{Proceedings of the European Conference on
  Computer Vision (ECCV)}, 2018, pp. 3--19.

\bibitem{li2018harmonious}
W.~Li, X.~Zhu, and S.~Gong, ``Harmonious attention network for person
  re-identification,'' in \emph{Proceedings of the IEEE Conference on Computer
  Vision and Pattern Recognition}, 2018, pp. 2285--2294.

\bibitem{paigwar2019attentional}
A.~Paigwar, O.~Erkent, C.~Wolf, and C.~Laugier, ``Attentional pointnet for
  3d-object detection in point clouds,'' in \emph{Proceedings of the IEEE
  Conference on Computer Vision and Pattern Recognition Workshops}, 2019, pp.
  0--0.

\bibitem{sun2018pointgrow}
Y.~Sun, Y.~Wang, Z.~Liu, J.~E. Siegel, and S.~E. Sarma, ``Pointgrow:
  Autoregressively learned point cloud generation with self-attention,''
  \emph{arXiv preprint arXiv:1810.05591}, 2018.

\bibitem{zhang2019pcan}
W.~Zhang and C.~Xiao, ``Pcan: 3d attention map learning using contextual
  information for point cloud based retrieval,'' in \emph{Proceedings of the
  IEEE Conference on Computer Vision and Pattern Recognition}, 2019, pp.
  12\,436--12\,445.

\bibitem{liu2019point2sequence}
X.~Liu, Z.~Han, Y.-S. Liu, and M.~Zwicker, ``Point2sequence: Learning the shape
  representation of 3d point clouds with an attention-based sequence to
  sequence network,'' in \emph{Proceedings of the AAAI Conference on Artificial
  Intelligence}, vol.~33, 2019, pp. 8778--8785.

\bibitem{zhiheng2019pyramnet}
K.~Zhiheng and L.~Ning, ``Pyramnet: Point cloud pyramid attention network and
  graph embedding module for classification and segmentation,'' \emph{arXiv
  preprint arXiv:1906.03299}, 2019.

\bibitem{wang2019graph}
L.~Wang, Y.~Huang, Y.~Hou, S.~Zhang, and J.~Shan, ``Graph attention convolution
  for point cloud semantic segmentation,'' in \emph{Proceedings of the IEEE
  Conference on Computer Vision and Pattern Recognition}, 2019, pp.
  10\,296--10\,305.

\bibitem{chen2019gapnet}
C.~Chen, L.~Z. Fragonara, and A.~Tsourdos, ``Gapnet: Graph attention based
  point neural network for exploiting local feature of point cloud,''
  \emph{arXiv preprint arXiv:1905.08705}, 2019.

\bibitem{velivckovic2017graph}
P.~Veli{\v{c}}kovi{\'c}, G.~Cucurull, A.~Casanova, A.~Romero, P.~Lio, and
  Y.~Bengio, ``Graph attention networks,'' \emph{arXiv preprint
  arXiv:1710.10903}, 2017.

\bibitem{badrinarayanan2017segnet}
V.~Badrinarayanan, A.~Kendall, and R.~Cipolla, ``Segnet: A deep convolutional
  encoder-decoder architecture for image segmentation,'' \emph{IEEE
  transactions on pattern analysis and machine intelligence}, vol.~39, no.~12,
  pp. 2481--2495, 2017.

\bibitem{loshchilov2016sgdr}
I.~Loshchilov and F.~Hutter, ``Sgdr: Stochastic gradient descent with warm
  restarts,'' \emph{arXiv preprint arXiv:1608.03983}, 2016.

\bibitem{simonovsky2017dynamic}
M.~Simonovsky and N.~Komodakis, ``Dynamic edge-conditioned filters in
  convolutional neural networks on graphs,'' in \emph{Proceedings of the IEEE
  conference on computer vision and pattern recognition}, 2017, pp. 3693--3702.

\bibitem{klokov2017escape}
R.~Klokov and V.~Lempitsky, ``Escape from cells: Deep kd-networks for the
  recognition of 3d point cloud models,'' in \emph{Proceedings of the IEEE
  International Conference on Computer Vision}, 2017, pp. 863--872.

\bibitem{atzmon2018point}
M.~Atzmon, H.~Maron, and Y.~Lipman, ``Point convolutional neural networks by
  extension operators,'' \emph{arXiv preprint arXiv:1803.10091}, 2018.

\bibitem{Liu_2019_ICCV}
Y.~Liu, B.~Fan, G.~Meng, J.~Lu, S.~Xiang, and C.~Pan, ``Densepoint: Learning
  densely contextual representation for efficient point cloud processing,'' in
  \emph{The IEEE International Conference on Computer Vision (ICCV)}, October
  2019.

\bibitem{Thomas_2019_ICCV}
H.~Thomas, C.~R. Qi, J.-E. Deschaud, B.~Marcotegui, F.~Goulette, and L.~J.
  Guibas, ``Kpconv: Flexible and deformable convolution for point clouds,'' in
  \emph{The IEEE International Conference on Computer Vision (ICCV)}, October
  2019.

\bibitem{te2018rgcnn}
G.~Te, W.~Hu, A.~Zheng, and Z.~Guo, ``Rgcnn: Regularized graph cnn for point
  cloud segmentation,'' in \emph{Proceedings of the 26th ACM international
  conference on Multimedia}, 2018, pp. 746--754.

\bibitem{ben20183dmfv}
Y.~Ben-Shabat, M.~Lindenbaum, and A.~Fischer, ``3dmfv: Three-dimensional point
  cloud classification in real-time using convolutional neural networks,''
  \emph{IEEE Robotics and Automation Letters}, vol.~3, no.~4, pp. 3145--3152,
  2018.

\bibitem{wu20153d}
Z.~Wu, S.~Song, A.~Khosla, F.~Yu, L.~Zhang, X.~Tang, and J.~Xiao, ``3d
  shapenets: A deep representation for volumetric shapes,'' in
  \emph{Proceedings of the IEEE conference on computer vision and pattern
  recognition}, 2015, pp. 1912--1920.

\bibitem{scanWeb}
HKUST-VGD, ``3d scene understanding benchmark,''
  \url{https://hkust-vgd.github.io/benchmark/}, 2020, accessed: 2020-07-13.

\bibitem{zhao2018psanet}
H.~Zhao, Y.~Zhang, S.~Liu, J.~Shi, C.~C. Loy, D.~Lin, and J.~Jia, ``Psanet:
  Point-wise spatial attention network for scene parsing,'' in
  \emph{Proceedings of the European Conference on Computer Vision (ECCV)},
  2018, pp. 267--283.

\bibitem{huang2019ccnet}
Z.~Huang, X.~Wang, L.~Huang, C.~Huang, Y.~Wei, and W.~Liu, ``Ccnet: Criss-cross
  attention for semantic segmentation,'' in \emph{Proceedings of the IEEE/CVF
  International Conference on Computer Vision}, 2019, pp. 603--612.

\bibitem{yi2016scalable}
L.~Yi, V.~G. Kim, D.~Ceylan, I.~Shen, M.~Yan, H.~Su, C.~Lu, Q.~Huang,
  A.~Sheffer, L.~Guibas \emph{et~al.}, ``A scalable active framework for region
  annotation in 3d shape collections,'' \emph{ACM Transactions on Graphics
  (TOG)}, vol.~35, no.~6, p. 210, 2016.

\bibitem{chang2015shapenet}
A.~X. Chang, T.~Funkhouser, L.~Guibas, P.~Hanrahan, Q.~Huang, Z.~Li,
  S.~Savarese, M.~Savva, S.~Song, H.~Su \emph{et~al.}, ``Shapenet: An
  information-rich 3d model repository,'' \emph{arXiv preprint
  arXiv:1512.03012}, 2015.

\bibitem{song2015sun}
S.~Song, S.~P. Lichtenberg, and J.~Xiao, ``Sun rgb-d: A rgb-d scene
  understanding benchmark suite,'' in \emph{Proceedings of the IEEE conference
  on computer vision and pattern recognition}, 2015, pp. 567--576.

\bibitem{qi2019deep}
C.~R. Qi, O.~Litany, K.~He, and L.~J. Guibas, ``Deep hough voting for 3d object
  detection in point clouds,'' in \emph{Proceedings of the IEEE/CVF
  International Conference on Computer Vision}, 2019, pp. 9277--9286.

\bibitem{long2015fully}
J.~Long, E.~Shelhamer, and T.~Darrell, ``Fully convolutional networks for
  semantic segmentation,'' in \emph{Proceedings of the IEEE conference on
  computer vision and pattern recognition}, 2015, pp. 3431--3440.

\end{thebibliography}
\end{document}